A Cost Effective Solution for Road Crack Inspection using Cameras and Deep Neural Networks

[1]Qipei Mei and [2]Mustafa Gül

[1] Ph.D. Student

[2]Associate Professor (Corresponding Author: mustafa.gul@ualberta.ca)

Department of Civil and Environmental Engineering, University of Alberta
Edmonton, Alberta, T6G 2W2, Canada

***Abstract:*** Automatic crack detection on pavement surfaces is an important research field in the scope of developing an intelligent transportation infrastructure system. In this paper, a cost effective solution for road crack inspection by mounting commercial grade sport camera, GoPro, on the rear of the moving vehicle is introduced. Also, a novel method called ConnCrack combining conditional Wasserstein generative adversarial network and connectivity maps is proposed for road crack detection. In this method, a 121-layer densely connected neural network with deconvolution layers for multi-level feature fusion is used as generator, and a 5-layer fully convolutional network is used as discriminator. To overcome the scattered output issue related to deconvolution layers, connectivity maps are introduced to represent the crack information within the proposed ConnCrack. The proposed method is tested on a publicly available dataset as well our collected data. The results show that the proposed method achieves state-of-the-art performance compared with other existing methods in terms of precision, recall and F1 score.

## *1. Introduction*

Cracks on road surfaces are early signs for potential damage in the pavements and in the supporting structures. They serve as a good indicator to assess the current condition of the transportation infrastructure. Defects in road surfaces may delay traffic and even cause safety issues if they are severe. In addition, our road infrastructure must be improved significantly to support the autonomous vehicles of the future in the scope of smart cities. The current common practice in road surface survey is mainly based on manual inspection, which has limitations like high costs and low efficiency. Such cracks may be present for a considerable amount of time before they are repaired.

In this context, the automation of crack detection on road surface is invaluable and a vast amount of research has been conducted in this field [1-4]. Efforts have been made to design specialized vehicles with professional sensors such as laser scanner for automated pavement crack detection. However, these vehicles are expensive, and, as a result, the inspection efficiency is low, and the inspection interval is usually large due to the limit of budget. As an example, the Ohio Department of Transportation spent 1,179,000 USD on purchasing such a specialized vehicle and the yearly maintenance fee for this vehicle is 70,000 USD [5].

Commercial grade cameras are of interests to researchers for developing automated crack/defect detection technologies because of their low cost and easy accessibility [6]. However, some early works indicated that it is challenging to use images captured by cameras for road crack detection [7, 8]. Since cameras do not provide depth information unlike laser sensors, they are more susceptible to environmental effects such as illumination changes or road texture changes. In addition, the irregular shape of cracks and the existence of noise and stains on the roads make it difficult to find a general approach that works for different situations.

Owing to the rapid development of deep learning methods in computer vision area, crack detection using camera captured images has become more feasible. Object detection algorithms, such as regions-based convolutional neural network (R-CNN) [9], Faster R-CNN [10] and you only look only (YOLO) [11], have achieved performances comparable to humans in complex situations. Researchers have made attempts to apply various deep learning algorithms to crack detection [3, 12, 13]. However, since cracks do not have a certain shape and usually have extremely large aspect ratios, the crack detection task is very different than other regular object detection tasks.

In this paper, a road inspection solution by mounting commercial-grade sport camera at the back of a moving vehicle is investigated. Data are collected from the road tests to create a dataset consisting of 600 images with different cracks. A novel algorithm based on deep neural network called ConnCrack is proposed to detect the cracks at pixel level. The contribution of this paper mainly includes: 1) The feasibility of using commercial grade sport camera mounted at the rear of a car operating at traffic speed is verified, and a new challenging pixel-level annotated dataset is introduced to consider the real life situation [14]; 2) a novel method combining conditional Wasserstein generative adversarial network and connectivity maps is developed for pixel level crack detection.

This paper is organized in the following structure. Section 2 will review some related work for image-based crack detection. Section 3 will describe the details of the experimental setup. In section 4, the deep learning-based algorithm will be explained. Then, results and analysis as well as conclusions will be presented at last in sections 5 and 6.

## *2. Related work*

### *2.1. Rule-based Techniques*

In general, there are three major paths for crack detection utilizing images, rule based, machine learning based and deep learning-based methods. In rule-based methods, different combinations of filters and image processing techniques are applied to identify the cracks in images.

Gavilán et al. [15] proposed an approach combining a series of image processing techniques. First, the image was preprocessed to enhance the linear features, and non-crack feature detection was conducted to eliminate confusing area like joints or filled cracks on pavements. Then, a seed-based approach combining multiple directional non-minimum suppression with symmetry check was proposed. Zou et al. [16] developed a three step method called CrackTree. In their method, the shadow was first removed using a geodesic based algorithm. Then, a probability map was created based on tensor voting. Finally, recursive tree-edge pruning was conducted on the minimum spanning tree generated on the probability map to identify cracks. Amhaz et al. [17] introduced an improved minimal path selection algorithms with a refined artifact filtering step so that the thickness of the crack pattern can be estimated. Their approach showed superior performance than another 5 existing methods in their paper.

Overall, the major advantage of rule-based methods is that neither annotation nor training process is required, so it is easier to implement the methods and verify the performance. The biggest disadvantage of this kind of methods that most of the features are handcrafted on some given datasets. In general, they cannot consider all the variation in real life images, and in most cases one method may work in one certain situation but will not work in another.

### *2.2. Machine Learning-based Techniques*

Realizing the complexity in texture of pavement surfaces, variation in the illumination and the irregularity in shapes of the cracks, researchers tend to seek machine learning based algorithms for crack detection starting last decade. Comparing with traditional rule-based techniques, machine learning based algorithms can implicitly consider a variety of the factors that could affect the appearance of cracks in the training process.

Hu et al. [18] treated the pavement as texture surface and cracks as inhomogeneity, and used texture analysis and shape descriptors to extract features. Support vector machines were used to classify whether a sub-region was crack or non-crack. Mathavan et al. [19] applied an unsupervised learning algorithm called self-organizing map to the crack images. Texture and color properties were integrated within the self-organizing map to distinguish cracks from background. Shi et al. [20] proposed a crack detection method based on random structured forests. In their method, integral channel features were introduced to learn the crack tokens with structured information. Then, random structured forest was applied to process the tokens and find the cracks.

### *2.3. Deep Learning-based Techniques*

Deep learning, as a branch of machine learning, has drawn much attention in last few years due to its superior performance in object detection and semantic segmentation [11, 21]. They were first time applied to crack detection task in 2016 [22]. In general, deep learning-based crack detection methods can be categorized into two groups, i.e., region based and pixel-based methods.

The region-based method is less computationally intensive and has been studied by a number of researchers. Cha et al. [23] developed a CNN and applied it to 40,000 regions with a resolution of 256×256 pixels for training. The algorithm can detect cracks by classifying each region separately. Gopalakrishnan et al. [24] utilized a per-trained deep CNN model and applied transfer learning to hot-mix asphalt and Portland cement concrete pavement images. Their algorithm can identify whether an image has crack or not in it. Hoang et al. [25] compared a CNN model with metaheuristic optimized edge detection algorithm. They showed that the performance of CNN was significantly better than edge detector.

However, the region-based methods can only provide information about the existence of cracks and rough shape and location depending on the size of regions. The value of crack detection decreases if the accurate pattern and location of the cracks cannot be given. To resolve this issue, pixel-level crack detection are studied. Ni et al. [26] developed a method comprising two deep neural networks. The first neural network was called GoogLeNet which served as a feature extractor. Then, a second neural network including bilinear deconvolution layer and eltwise operation layer were used for pixel-level crack detection. Fei et al. [3] designed a deep neural network consisting of a preprocess layer, eight convolutional layers, and one output layer. With invariant spatial size through all layers, the method can achieve pixel level crack detection. Yang et al. [13] utilized a fully convolutional neural network (FCN) to realize the pixel level detection. Through the encoder and decoder process, the output was guaranteed to be the same size as input. Therefore, the prediction was included in the output probability map.

The deep learning-based algorithms have shown great potential in solving crack detection problems on pavement surface. However, there are still remaining challenges due to the issues such as inhomogeneity of cracks, complexity of illumination conditions, and similarity of appearance between cracks and pavement textures. In the authors' opinion, one of the biggest restrictions that holds back the fast development of novel algorithms is the lack of high quality and challenging datasets with complete annotations. In above studies, the researchers either tested their methods on their own datasets [13, 23-26] or very simple publicly available datasets [3]. In this context, it is difficult to compare the performance among algorithms and for new researchers to test their methods. In this paper, a challenging dataset collected by our group will be introduced in the next section.

## *3. Experimental Setup and Data Collection*

In this study, field experiments were conducted with a GoPro Hero 7 Black mounted beside the license plate on the rear of a Honda Pilot 2017 (see Fig. 1) for data collection. It should be noted that this experiment has an initial goal to mimic the behavior of a backup camera in a vehicle. Current vehicles usually do not allow access to their backup camera systems easily. Therefore, the camera was placed at the same level and facing the same direction as the backup camera in this car to mimic its behavior. Data are collected based on this setting with the expectation that the conclusions drawn in this study could be useful in the future when the access to backup camera images becomes more practical.

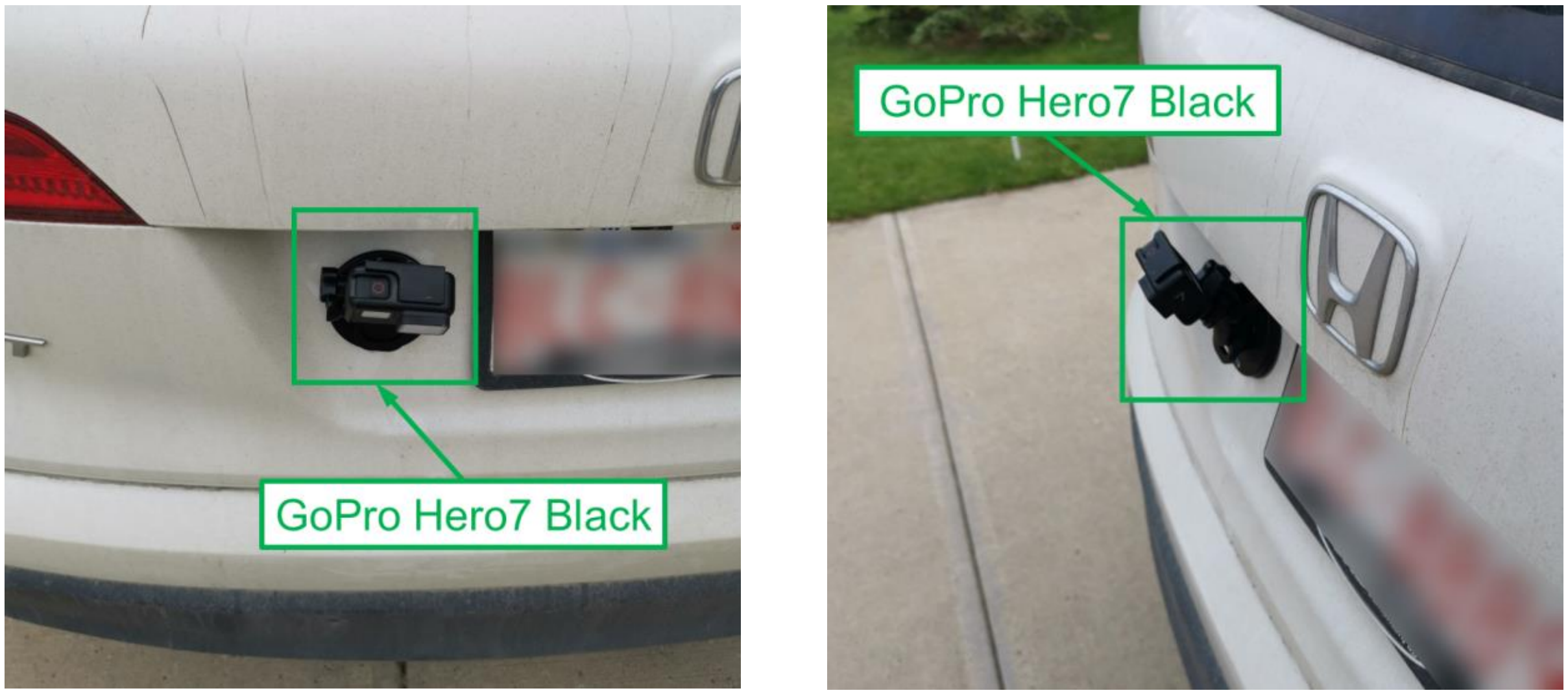

(a) front view (b) perspective view

Fig. 1. GoPro Hero 7 Black mounted at the rear of Honda Pilot 2017

In several previous studies [27, 28], the camera was mounted behind the windshield in the front of the car. As presented in Fig. 2, two configurations are illustrated and compared. In rear-mount configuration, the angle of camera is set to 45° to balance the spatial resolution and scanned area. In front-mount configuration, the camera is facing forward like in previous studies [27, 28]. In these two configurations, the spatial resolution defined as number of pixels in unit length can be calculated as in Eq. (1). The spatial resolution represents how much detail can be captured by the camera.

$$\rho = 1 / \left[ d\tan(\alpha + \Delta\theta + \theta / m) - d\tan(\alpha + \Delta\theta) \right] \tag{1}$$

where $d$ is the distance from the center of camera lens to the ground, $\Delta\theta$ is the angle from the bottom line of FOV, $\theta$ is the FOV, $m$ is the total number of pixels in vertical direction and $\alpha$ is the angle between bottom line of FOV and vertical line.

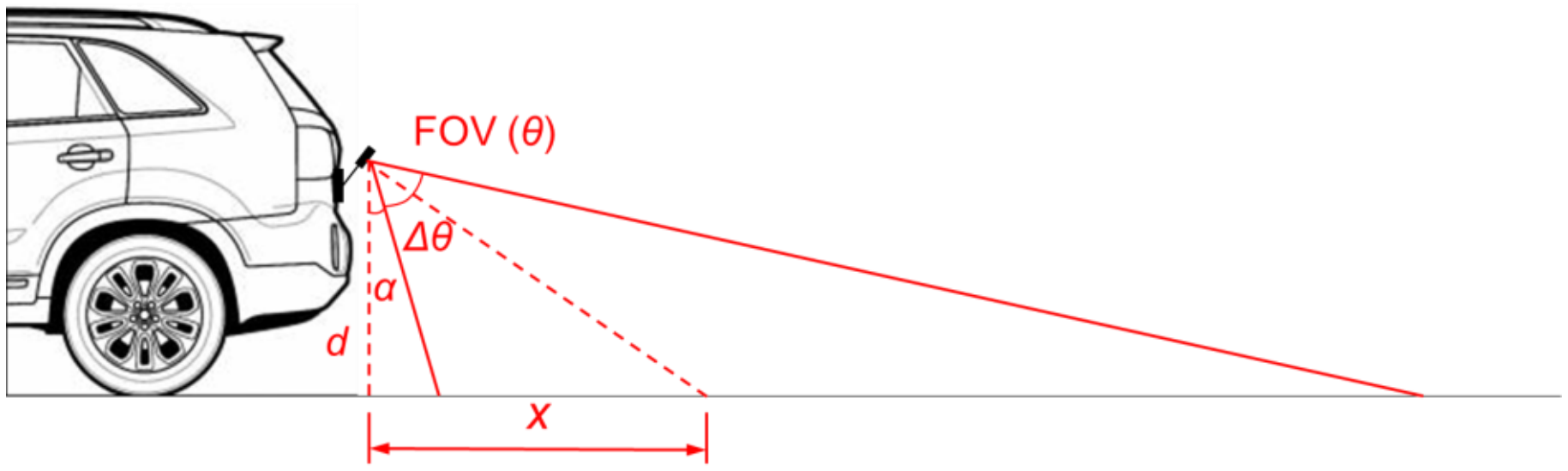

(a) Rear-mount (proposed)

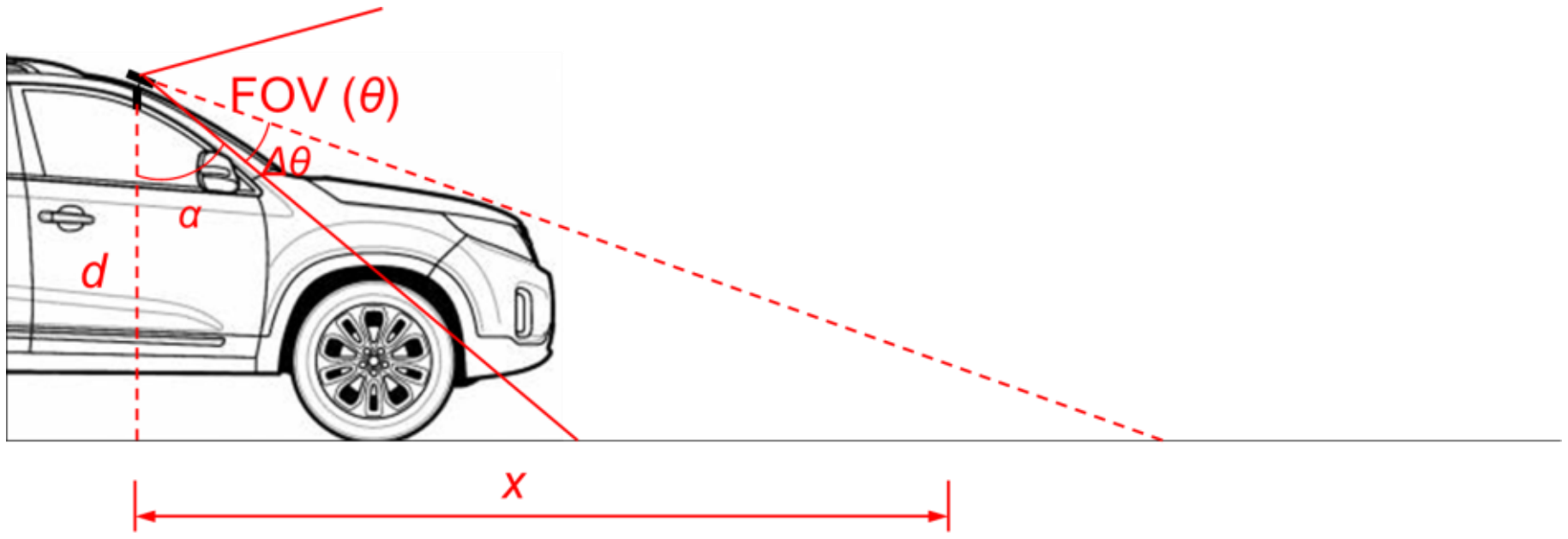

(b) Front-mount

Fig. 2. Comparison of two configurations (modified from [29])

Table 1 – Comparison between different mounting strategies

| Percentage of FOV, Δθ/θ | Spatial Resolution (pixel/cm) | |
|---|---|---|
| | Rear-mount | Front-mount |
| 0% | 8.62 | 1.93 (blocked) |
| 25% | 6.99 | 0.53 |
| 50% | 4.45 | 0.00 |
| 75% | 1.91 | N/A |
| 100% | 0.28 | N/A |

In this study, the GoPro Hero 7 black has a FOV of 69.5º. The image has a resolution of 1920×1080 pixels. Therefore, $\alpha$ for rear-mount configuration is 45°-69.5°/2=10.25° and for front-mount configuration is 90°-69.5°/2=55.25°. The vertical distance to the ground is 1.5 m for front-mount configuration and 1 m for rear-mount configuration. According to the above information, the parameters in Table 1 are calculated. In Table 1, the percentage of FOV is corresponding to percentage of image regarding the image bottom in vertical direction. For instance, $\Delta\theta/\theta$ of 50% means the centerline of the image in vertical direction. It is seen from the table that the spatial resolution decreases dramatically as the percentage of FOV increase, which is expected because the pavement is farther from the camera. Comparing these two configurations, we can see the front-mount configuration has significantly less spatial resolution than rear-mount configuration overall. This is because the front-mount camera is farther from the ground. Also, the 0% to 25% region is most likely to be blocked by the hood. Therefore, it is seen that rear-mount configuration to better utilize the GoPro camera.

To summarize, there are three main reason that we use a rear-mount configuration: 1) The windshield could reflect the light inside of the car and reduce the quality of the image in front-mount configuration. 2) The front camera is farther from the ground, a large part of its field of view (FOV) is blocked by the hood of the car. Therefore, the front-mount configuration sacrifices too much spatial resolution corresponding to our analysis above. 3) Our eventual goal is to directly use backup camera in vehicles for crack detection while driving. In this case, no external devices need to be installed in this case.

The data were collected while the vehicle was driving at traffic speeds (40 kph – 80 kph), and 240 fps frame rate and 1/3840 sec shutter speed was used for the camera. In total, about 3-hour videos were taken from different roads in Edmonton, Canada at different times over two months by our research group. Images were extracted every 6 frames. After discarding those without cracks, we created a dataset called EdmCrack600 which includes 600 images with full annotation at pixel level. The data aim to cover various factors one could encounter on the roads like different weather conditions, different illumination conditions, existence shadows from other objects, texture difference among difference pavement surfaces, etc., so no specific restrictions are applied during the collection process. The dataset, EdmCrack600, will be made public to benefit the community [14]. Some sample images are shown in Fig. 3. It is seen that the collected dataset is more difficult than most of the publicly available ones.

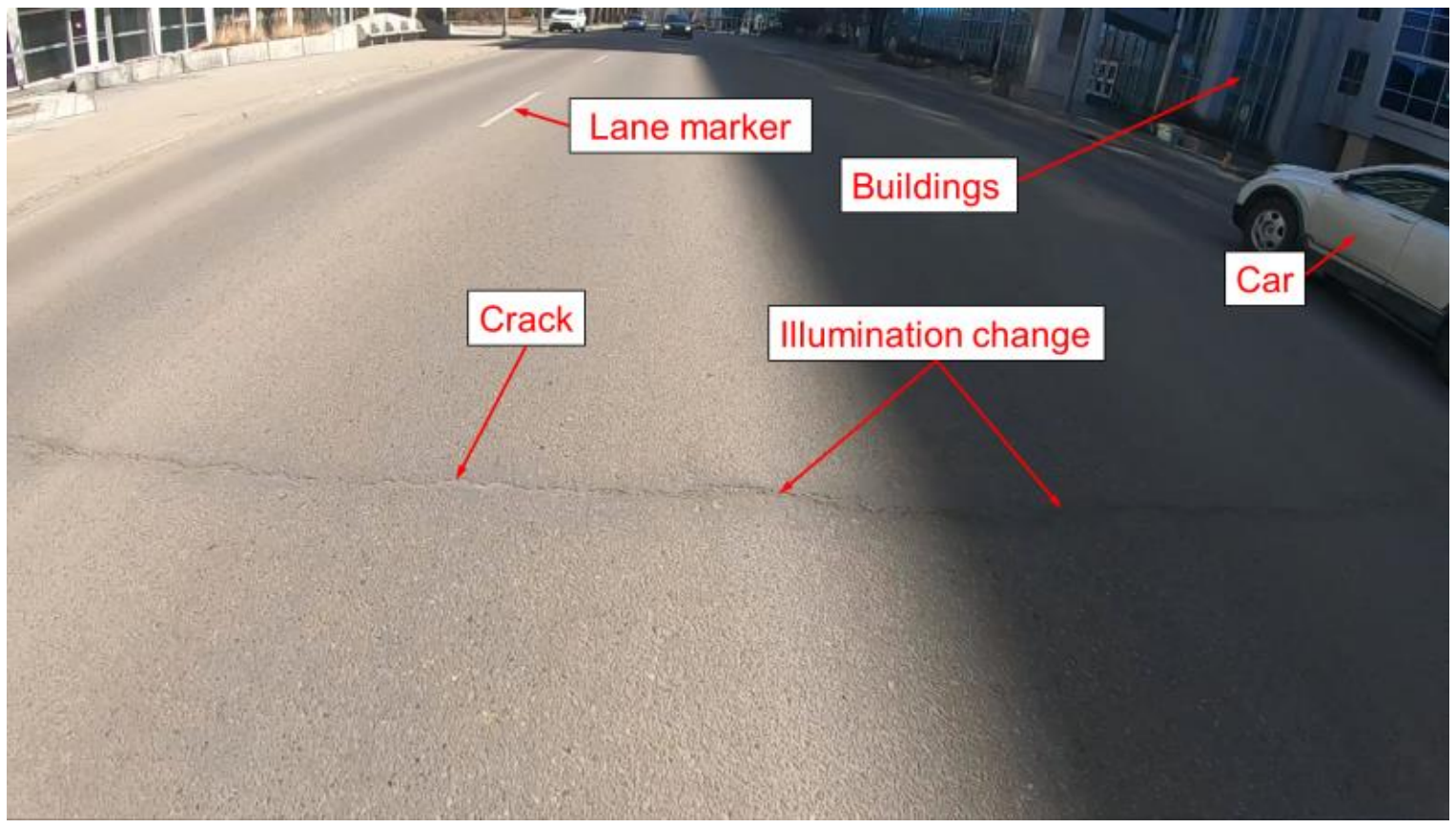

(a) Sample image 1

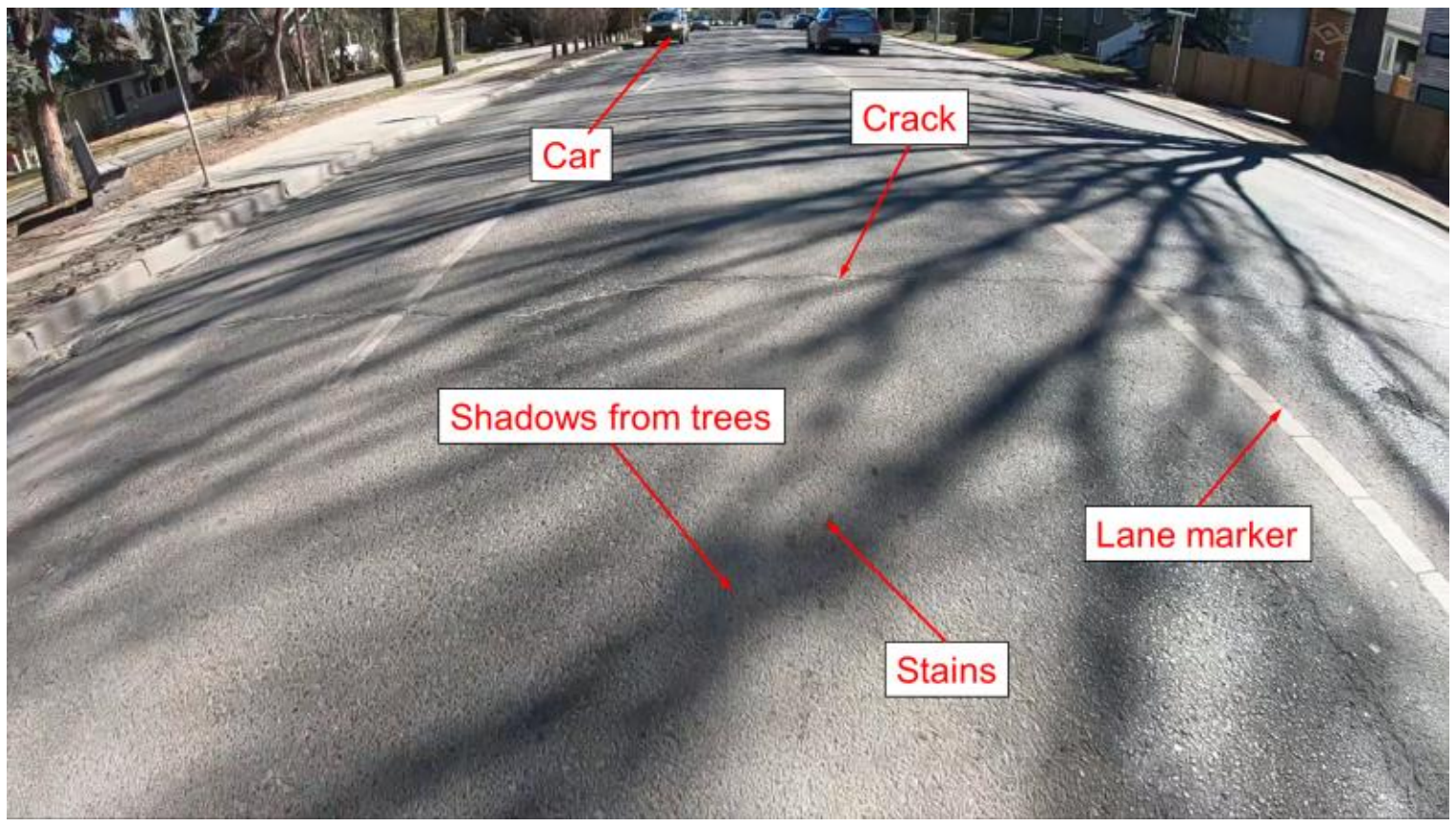

(b) Sample image 2

Fig. 3. Sample images from EdmCrack600 dataset

The publicly available datasets specifically designed to evaluate crack detection algorithms are limited. Furthermore, most of the datasets have been simplified comparing to the ones that could be encountered in real life. For example, some datasets control the light conditions [30], some manually exclude any disturbance and focus only on pavement surfaces using static images [17, 20, 31], and some were created for other algorithms and simply do not have enough images for deep learning [1, 17].

A comparison of this dataset and other publicly available dataset is given in Table 2. It is seen that only GAPs [30] and JapanRoad [28] datasets consist of more images than our dataset. However, those two datasets are not pixel-level annotated. The cracks in their datasets are annotated by bounding boxes. In authors' opinion, the bounding box is a not good way to annotate crack because of the irregular shape of cracks. Too many details will be lost if a rectangular bounding box is used to cover the cracks.

To the best of the authors' knowledge, our dataset, EdmCrack600, is the largest crack dataset so far which is annotated at pixel level. It is also a very challenging one because of all the factors that are taken into consideration during the data collection process. The challenges include: 1) Change of weather conditions; 2) Significant environmental effects and noise: shadows, occlusion, stains, texture difference, low contrast because of overexposure; 3) Blurring effect due to moving of the car and the poor lighting condition.

Table 2 – Comparison among different datasets

| Dataset | No. Images | Resolution | Device | Colored | Environmental effect* | Non-pavement region** | Pixel level annotation | Traffic speed | Extracted from video |
|---|---|---|---|---|---|---|---|---|---|
| CFD [20] | 118 | 480×320 | iPhone 5 | yes | yes | no | yes | no | no |
| Aigle-RN [17] | 38 | 991×462<br>311×462 | professional camera | no | no | no | yes | yes | no |
| Crack500 [31] | 500 | 2,000×1,500 | LG-H345 | yes | no | no | yes | no | no |
| GAPs [32] | 1969 | 1920×1080 | professional camera | no | no | no | no | yes | yes |
| Cracktree200 [16] | 206 | 800×600 | unknown | yes | yes | no | yes | no | no |
| GaMM [1] | 42 | 768×512<br>1920×480 | professional camera | no | yes | no | yes | yes | yes |
| CrackIT [33] | 84 | 1536×2048 | optical device | yes | no | no | yes | unknown | no |
| JapanRoad [21, 28] | 9,053 | 600×600 | LG-5X | yes | yes | yes | no | yes | no |
| EdmCrack600 (current study) | 600 | 1920×1080 | GoPro 7 | yes | yes | yes | yes | yes | yes |

*Environmental effect includes shadows, occlusions, low contrast, noise, etc.

**non-pavement region means the region of image that does not belong to pavement, such as cars, houses, sky.

# *4. Methodology*

## *4.1. Overall Procedure*

A novel deep learning-based algorithm called ConnCrack is proposed in this paper. The overall procedure of ConnCrack is described in Fig. 4. The method is developed on the basis of a conditional Wasserstein generative adversarial network (cWGAN), and connectivity maps are introduced to resolve scattered output due to deconvolution layers. The method consists of two neural networks which are termed as generator and discriminator. In this setting, the generator outputs connectivity maps for the identification of cracks, while the discriminator checks if the connectivity maps are ground truth ("real") or prediction ("fake"). Two networks are trained alternately to reach a Nash equilibrium after convergence [34].

In the generator, color image patches are taken as input, and a DenseNet121 with deconvolution layers for multiple-level feature fusion is applied. Unlike other deep learning-based crack detection methods, the generator outputs 8 connectivity maps instead of a binary probability mask. The reason and the advantages of this innovation will be explained in following sections. Then, the connectivity maps and the original patch will be fed into the discriminator to check if this is a "fake" or "real" output.

In this paper, the ConnCrack will be trained on patches which are subregions of the original large images to overcome the issues related to insufficient training data. The crack detection of the whole image will be integrated from the results coming from the patches. A post processing technique including a standard depth first search (DFS) algorithm to find connected components and to threshold out connected components with a small number of pixels is applied to the output the generator. The reason for this processing is because the cracks are usually connected components with a large number of pixels, but noise has much fewer connected pixels.

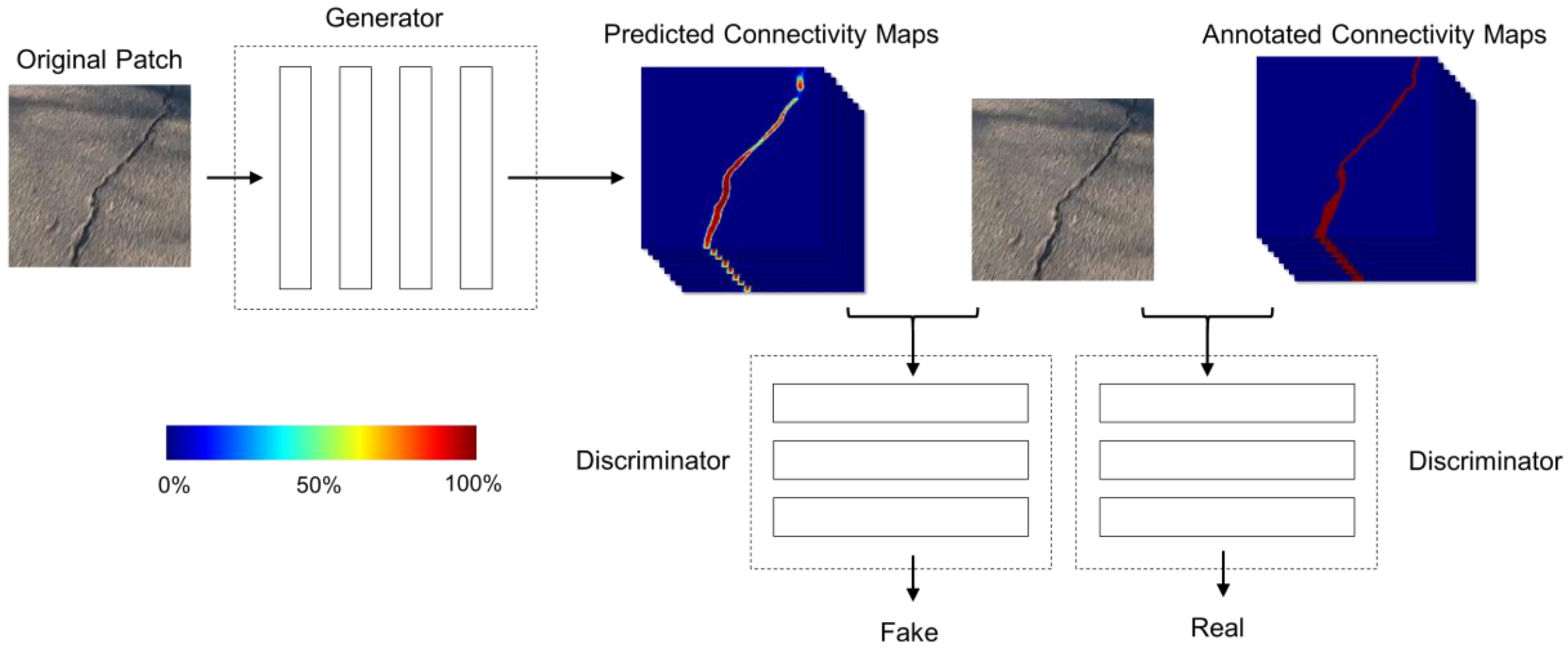

Fig. 4. Overview of the proposed method

## *4.2. Connectivity Maps*

In this paper, deconvolution layers are used for upsampling and pixel level identification similar to some other studies [13, 27] for computational efficiency. However, we realize that deconvolution layers are likely to generate scattered output (see Fig. 5(b)), i.e. the crack segments are not strictly connected. This is due to the mechanism of deconvolution layers where the predicted label of a pixel is solely dependent on the pixel values of a local region in original patch but is not explicitly related to the predicted labels of its neighboring pixels. Some studies suggested morphological operations, i.e., dilation and erosion, to resolve this issue [20]. However, as shown in Fig. 5(c) and (d), the performance is highly dependent on the selection of the size of morphological operations. If the size is too small, the gaps are not fully filled. If the size is too large, unnecessary parts will be considered as cracks.

This issue comes from the definition of cross entropy loss function currently used in many deep neural networks for crack detection [4, 27]. Taking Fig. 6 as an example, the crack pixels are labelled as 1 and the non-crack pixels are labelled as 0 in the ground truth. If the neural network mistakenly predicts one pixel within crack as 0, it is not different than predicting a non-crack as 1 in terms of loss function. However, in reality, an isolated wrong prediction is easier to fix than scattered prediction in crack segments.

To resolve this issue, we transform the crack detection into a connectivity problem inspired by [35]. Starting from the ground truth binary mask, each pixel should have 8 neighboring pixels. We generate 8 connectivity maps to reflect the relationship between a pixel and its 8 neighbors. As presented in Fig. 7, a regular ground truth binary crack mask is converted to 8 connectivity maps. For instance, one element in A2 connectivity map is 1 only if the corresponding element in ground truth binary mask is 1 and its left neighbor is 1 as well. During the training process, the ground truth connectivity maps are compared with predicted connectivity maps as one source to update the weights of the deep neural networks. The loss function based on the connectivity maps which is termed as $L_{content}$ could be written as Eq. (2) below.

$$
\begin{aligned}
L_{\text{content}}(G) &= E_{x,y}\left[-y\log G(x)-(1-y)\log(1-G(x)\right] \\
&= \sum_{k=1}^{8}\sum_{i,j\in\text{image}}\begin{bmatrix}-y_{A_k}(i,j)\log\hat{y}_{A_k}(i,j) \\ -(1-y_{A_k}(i,j))\log(1-\hat{y}_{A_k}(i,j))\end{bmatrix}
\end{aligned}
\quad (2)
$$

where $G$ represents the generator. It takes $x$ as input and generates $G(x)$. The true label (ground truth connectivity maps) of input $x$ is termed as $y$. Also, at pixel level, $y_{A_k}(i,j)$ is the true label of a pixel at $i$ and $j$ in the connectivity map $A_k$. And $\hat{y}_{A_k}(i,j)$ is the predicted label for the corresponding pixel.

With the help of connectivity maps, more weights will be given to the pixels within crack segments and less weights are given to isolated pixels. In this way, the predictions are forced to be connected to each other. As can be seen in Fig. 8, the performance of deep neural network trained with regular binary mask and our proposed connectivity maps are compared. The results based on connectivity maps are more robust and less scattered because the connectivity maps force the predictions to be connected.

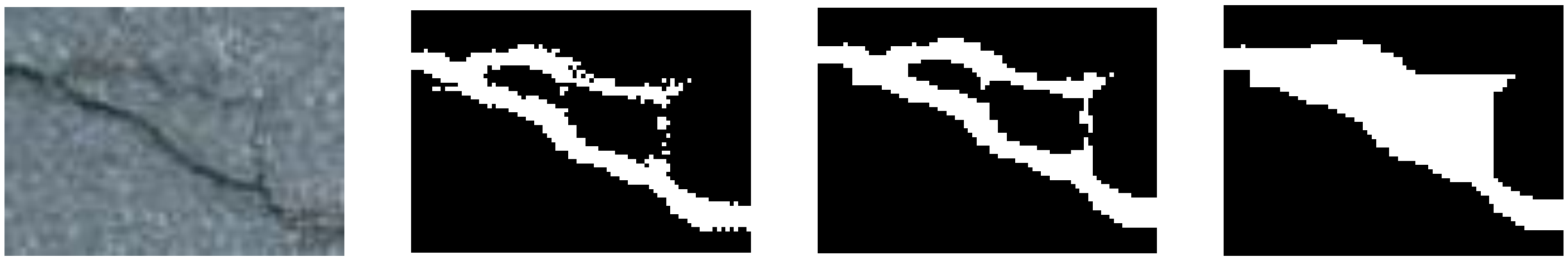

(a) (b) (c) (d)

Fig. 5. Issues with deconvolution layer output (a) original patch; (b) raw output; (c) after 3×3 morphological operations; (d) after 15×15 morphological operations

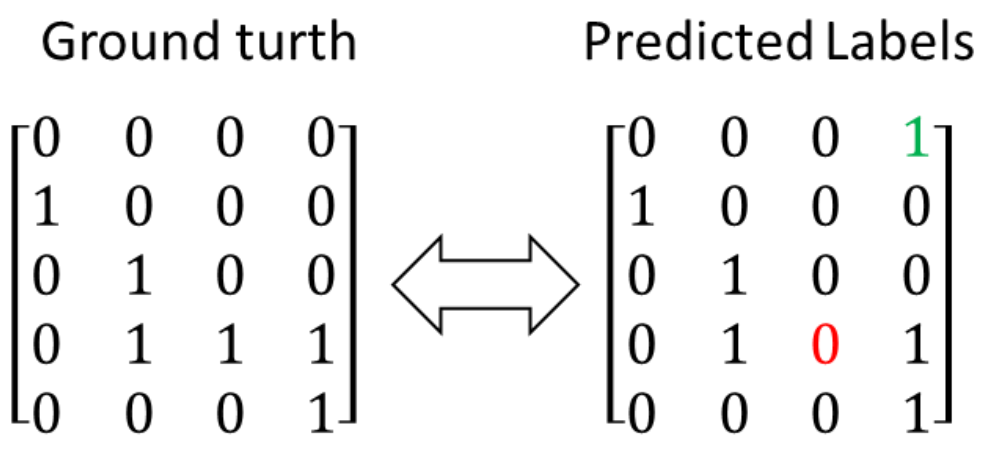

Fig. 6. An example of crack detection

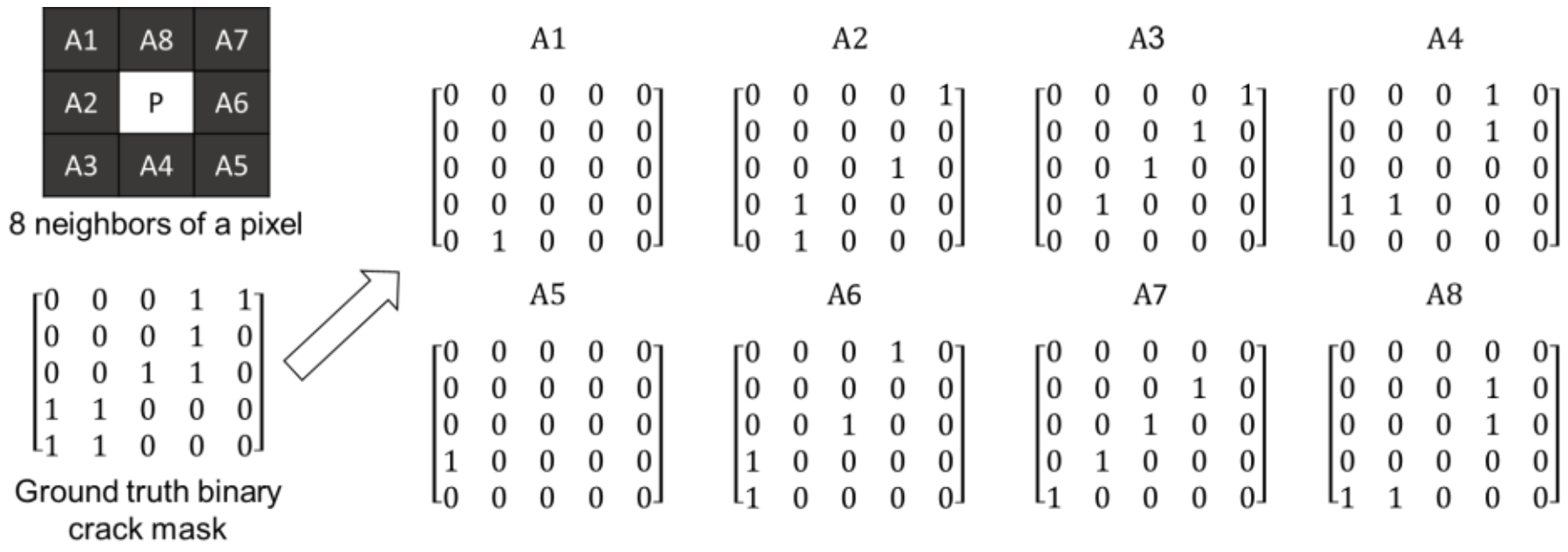

Fig. 7. Connectivity maps for crack annotation

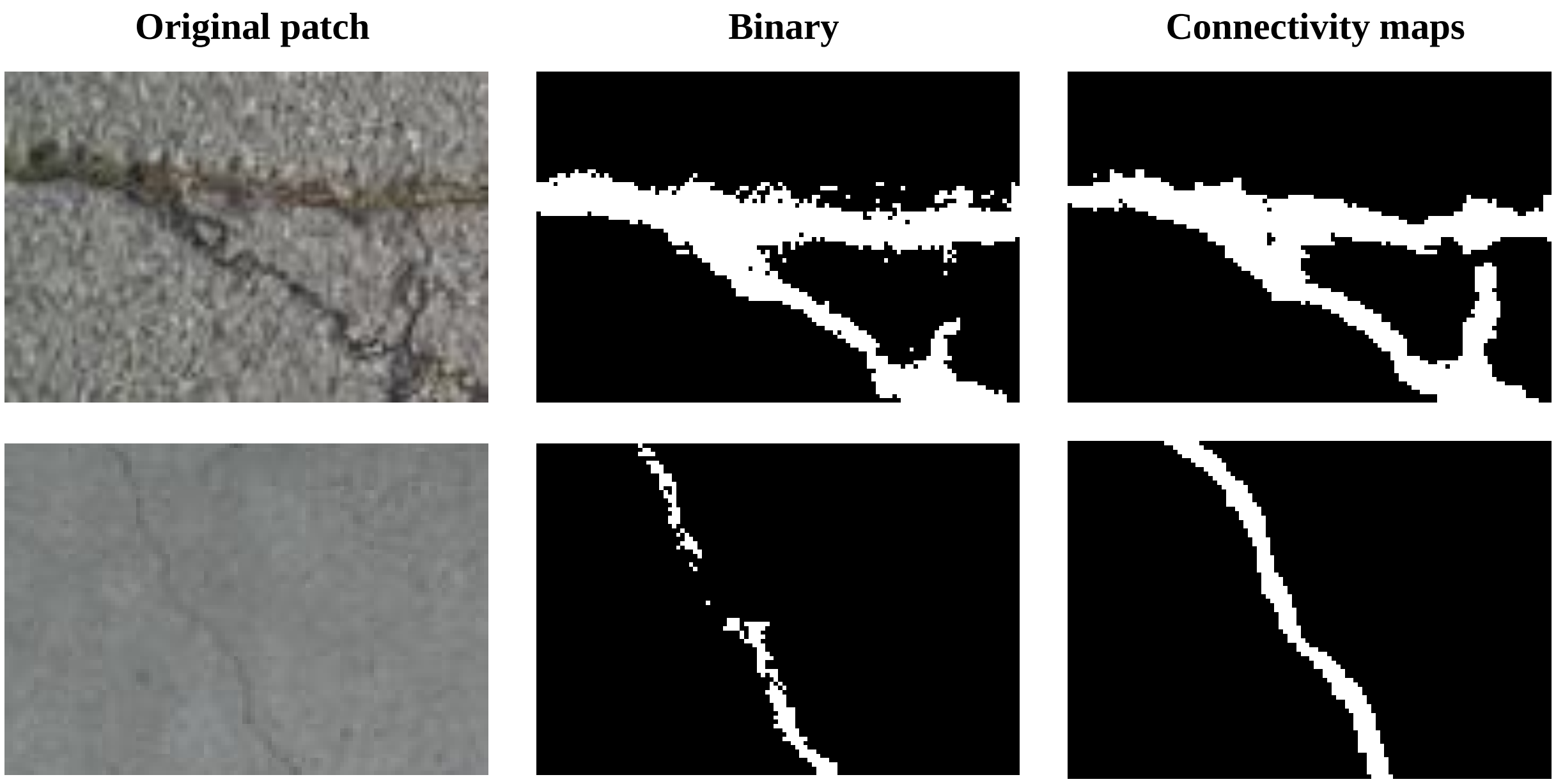

Fig. 8. Comparison between binary mask and connectivity maps

### *4.3. Generator*

The generator is the deep neural network for crack detection. In the proposed method, as shown in Fig. 9, a DenseNet121 [36] is used as feature extractor and 3 deconvolution layers are applied for multi-level feature fusion to generate target connectivity maps.

The DenseNet121 consists of a standalone convolutional layer, a max pooling layer, 4 dense blocks and 3 transition blocks. The convolutional layer was first proposed by LeCun [37], which is now widely used for computer vision problems. Similar to filters in traditional image processing techniques, a convolutional layer is applied to the input in a sliding window form. Unlike a fully connected layer, the sparsely connected neurons in a convolutional layer can lead to better efficiency and performance. Max pooling layer replaces the value of the input feature at a certain location with its neighboring features. It can reduce the size of features and make the features invariant to small translations.

One characteristic of DenseNet121 that distinguishes it from other deep neural networks is the application of the dense block. A dense block consists of a number of convolutional layers which are densely connected with each other in a feed-forward fashion. A 1×1 convolutional layer and a 3×3 convolutional layer form a basic component in a dense block. Each dense block has multiple such components, and each component is directly connected with all following basic components within this block using skip connections except the mainstream chain-like connections. In DenseNet121, the dense blocks 1, 2, 3 and 4 (see Fig. 9) have 6, 12, 24 and 16 basic components, respectively.

The dense block does not change the height and width of the features. To follow an encoder-decoder schema for pixel level crack identification, transition blocks are applied to reduce the size of features. A transition block composes of a 1×1 convolutional layer and a 2×2 average pooling layer with a stride of 2. The reduction of size is achieved by the average pooling layer in the transition block.

Deconvolution layers are applied to fuse features from multiple levels so that the predicted connectivity maps have the same height and width of the original patch. Unlike traditional upsampling techniques, such as bilinear and bicubic interpolations which have predefined parameters, the parameters for upsampling in deconvolution layers are determined during the training process. The deconvolution layers were first time used for upsampling in semantic segmentation by Long et al. [38]

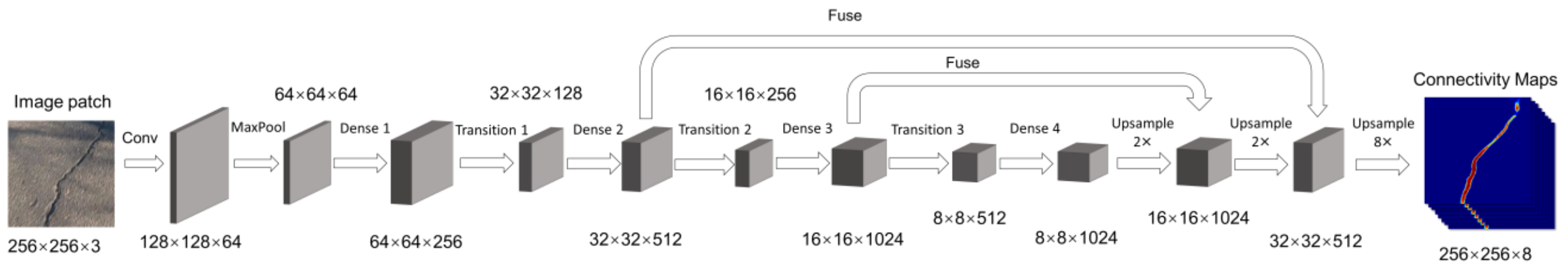

Fig. 9. Details of the generator

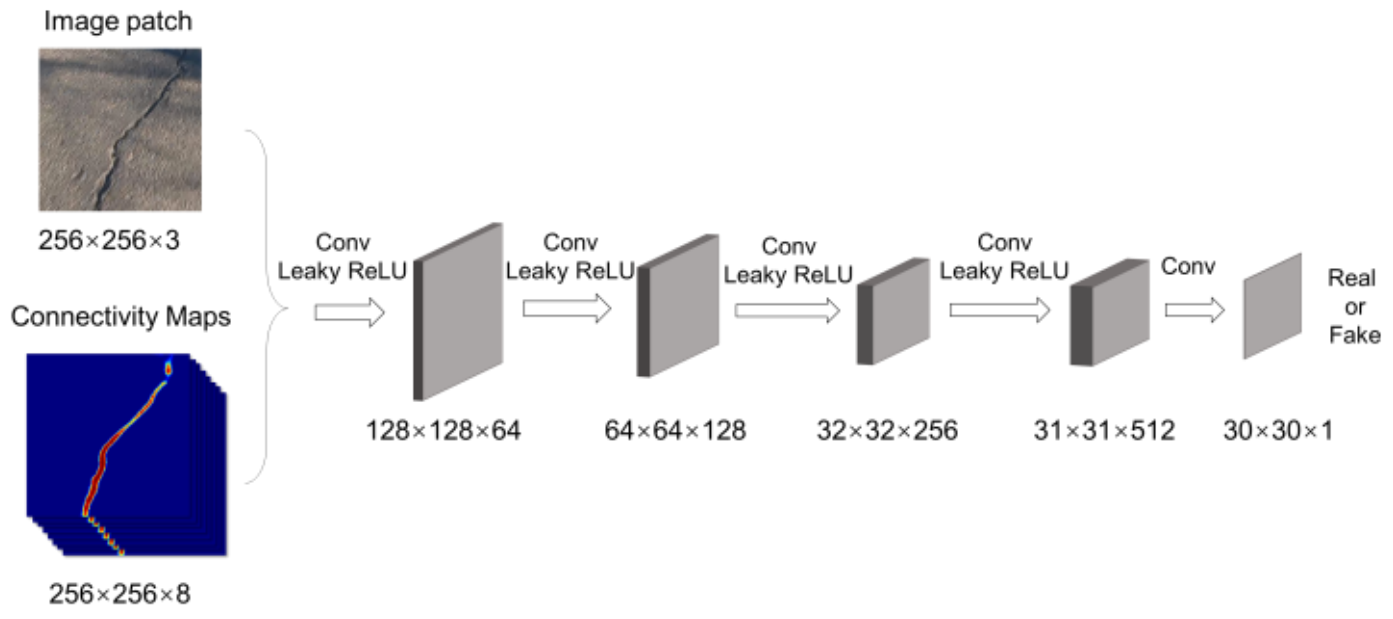

Fig. 10. Details of the discriminator

### *4.4. Discriminator*

In the ConnCrack, the architecture of the discriminator is presented in Fig. 10. It is 5-layer fully convolutional neural network. The original image patch and the corresponding connectivity maps are concatenated and passed through the discriminator. For the ground truth, the discriminator is expected to output labels as “real”. In contrast, it is expected to output “fake” when the predicted connectivity maps are used as input.

Similar to Pix2Pix [39] but different from traditional conditional generative adversarial networks (cGAN) [40], the proposed method uses a Markovian discriminator, where the output is not a single label but 30×30 labels representing “real” or “fake”. Each element of the 30×30 tensor corresponds to a small 70×70 patch, and it shows whether this patch is “real” or “fake”. These small patches are overlapped with each other. According to [39], the Markovian discriminator is better at capturing the high frequency part (details) of the image.

### *4.5. Loss Function*

The loss function used in the proposed method combines cWGAN loss and content loss. The loss function is given in Eq (3).

$$\begin{aligned} L_{cWGAN}(G,D) &= E_{x,y}\left[D(x,y)\right] - E_x\left[D(x,G(x))\right] \\ G^* &= \arg\min_G \max_D(\lambda L_{cWGAN}(G,D) + L_{\text{content}}(G)) \end{aligned} \tag{3}$$

where $x$ is the input patch, $y$ is the ground truth connectivity maps, $G$ is the generator, $D$ is the discriminator and $\lambda$ is the parameter adjusting the weights of $L_{\text{cWGAN}}(G, D)$ and $L_{\text{content}}(G)$.

Unlike traditional cGANs, the log functions are removed from $L_{cWGAN}(G, D)$ to achieve a Wasserstein distance following the suggestion from [41]. During the training process, the weights of the discriminator is clipped to a range [-*C*, *C*] to fulfill the requirement Lipschitz constraint [41] where *C* is a constant. Also, similar to [39], we add a content loss directly comparing with the output of the generator *G* with the ground truth. This could help the training process of the generator according to [42].

In the practical implementation, the discriminator *D* and generator *G* are trained alternatively. The generator *G* is trained to generate predicted connectivity maps ("fakes") that cannot be distinguished from ground truth ("reals") by discriminator *D*. In contrast, the discriminator *D* is trained to be better at distinguish the "reals" from "fakes". After the training is completed, the generator *G* will be used for crack detection, and the discriminator *D* can be discarded.

### *4.6. Evaluation*

Three metrics are used for the evaluation of the proposed method, i.e., precision, recall and F1 score. The formulae to calculate these metrics are given in Eq. (4).

$$\begin{aligned} \text{precision} &= \frac{TP}{TP+FP} \\ \text{recall} &= \frac{TP}{TP+FN} \\ \text{F1 score} &= \frac{2 \times \text{precision} \times \text{recall}}{\text{precision} + \text{recall}} \end{aligned} \tag{4}$$

In above equations, *TP* is true positive, *FP* is false positive, and *FN* is false negative. Following the definition given in [20], the *TP* is defined as the number of crack pixels that are within 5-pixel distance of a ground truth crack pixel. *FP* is the number of crack pixels that are beyond 5-pixel distance of a ground truth crack pixel. FN is the number of non-crack pixels that match the ground truth non-crack pixels.

## ***5. Analysis and Results***

### *5.1. Pretraining on ImageNet and CFD datasets*

From a number of previous studies, it is well accepted that pretraining on irrelevant large datasets in advance before tackling the task can help improve the performance of the deep learning-based algorithms [43]. This strategy is called transfer learning. In this paper, the proposed generator is first pretrained on a large object detection dataset called ImageNet [44]. It should be noted that the ImageNet dataset does not have a category related to pavement cracks.

Then, the whole proposed method is again pretrained and tested on a small crack dataset called CFD which was introduced by Shi et al. [20]. This dataset consists of 118 pavement images with resolution of 480 $\times$ 320 pixels. The images are taken by iPhone 5 with focus of 4 mm and aperture of *f/2.4*. In this paper, the dataset is split into 60%/40% for training and testing. More details of the dataset can be found in [20].

For the training and testing, the images are split into 128 $\times$ 128 patches and are then integrated to the original size after being processed by the ConnCrack. Both the learning rate and $\lambda$ are set to $1 \times 10^{-6}$ during the training. The training losses of generator and discriminator are presented in Fig. 11. For better visualization, a 5-element moving average is taken on all the curves. As can be seen in Fig. 11(a), the generator loss has two components, one comes from the cWGAN and the other comes from the content loss described in Eq. (2). It is seen that the content loss continuously decreases as the training proceeds. The cWGAN loss for generator first decreases and then increases since the discriminator has learned to distinguish the “fakes” from “reals”. Looking at Fig. 11(b), the loss for discriminator is low at the beginning but increases afterwards. This is because initially the generator is not well trained, and the discriminator can easily distinguish the generated output from the ground truth. However, as the training proceeds, the generator can output predictions that are more difficult to distinguish. In this context, the loss for discriminator starts to increase.

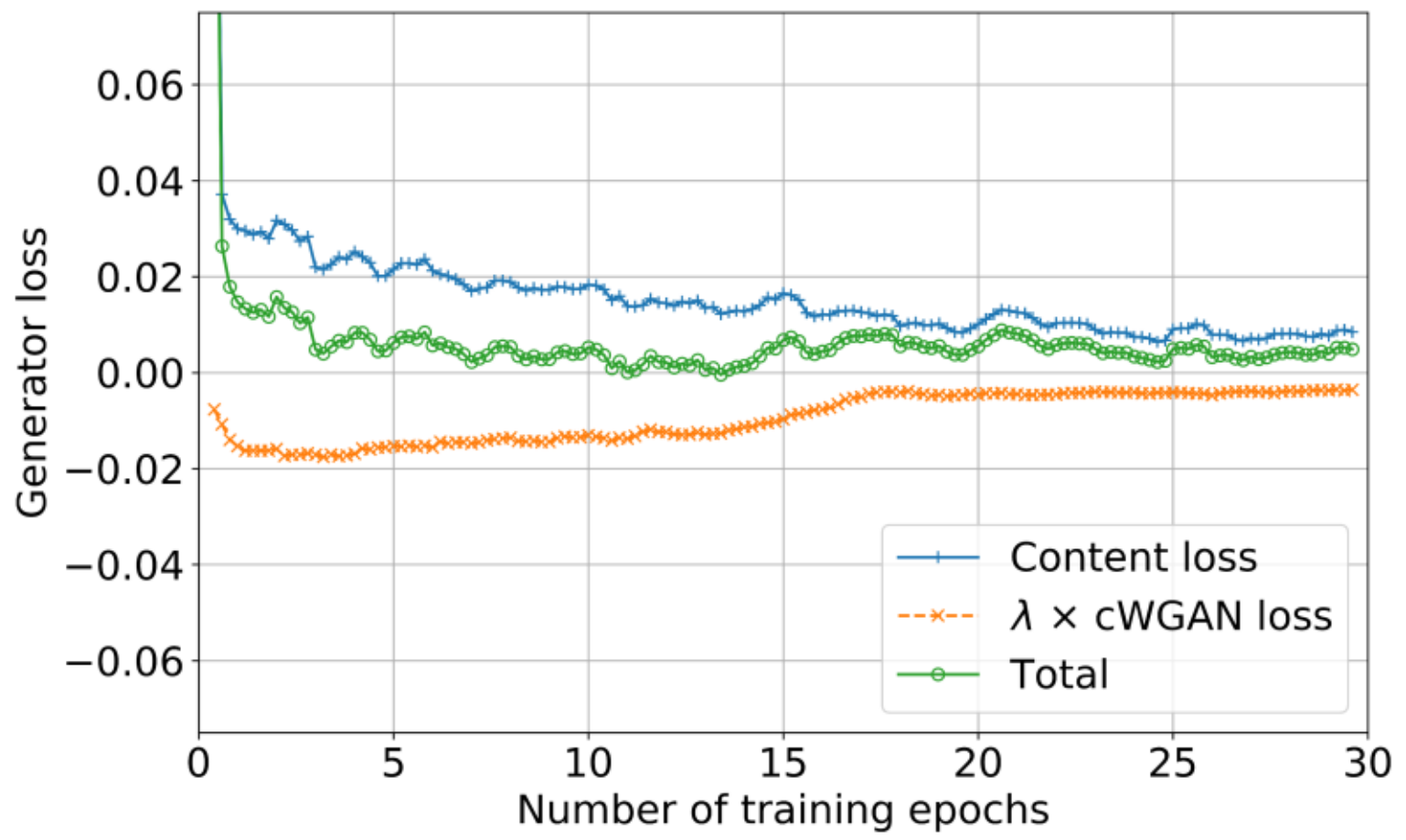

(a) Generator loss

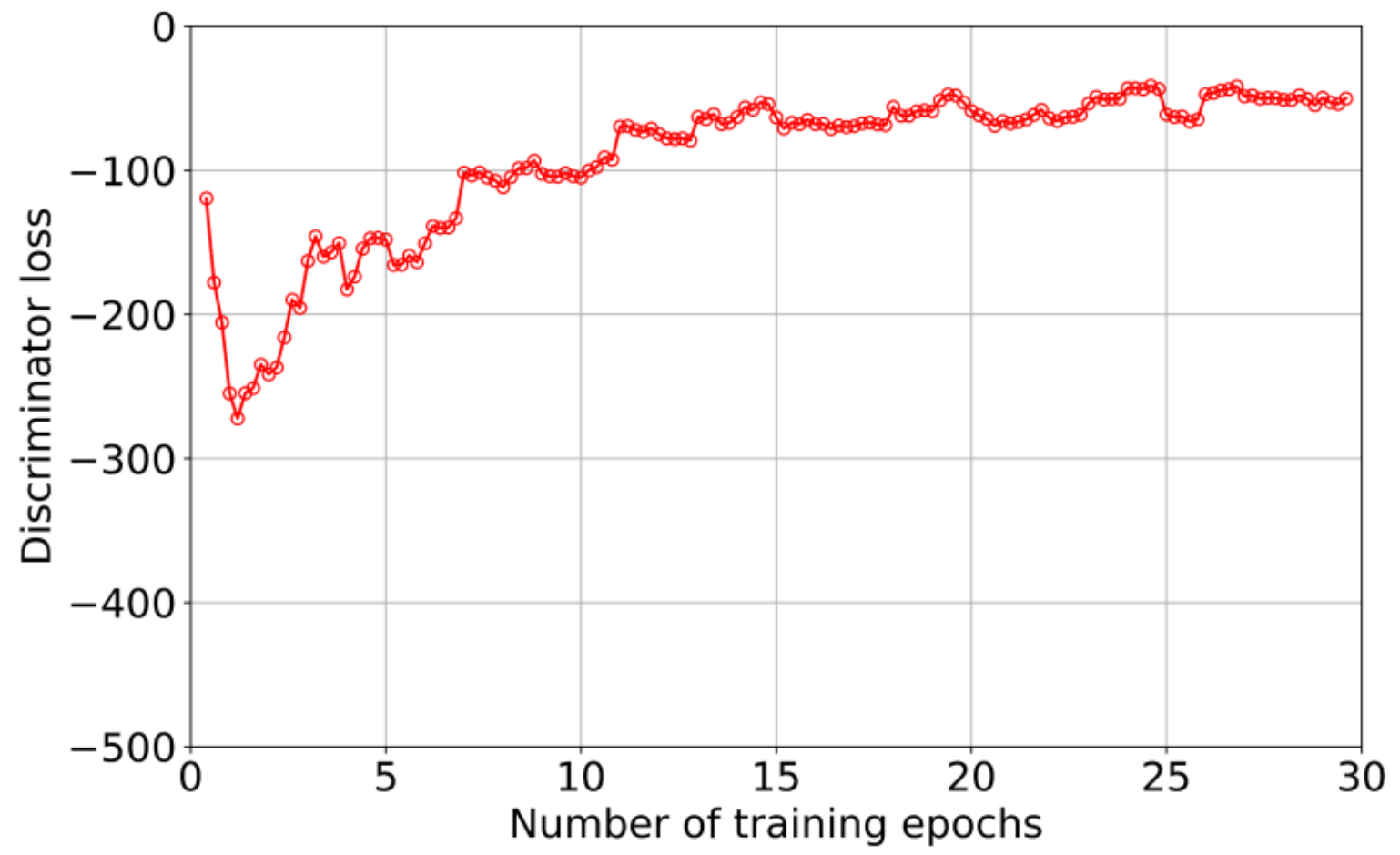

(b) Discriminator loss

Fig. 11. Losses of the proposed method

Some sample images along with the ground truth and prediction are presented in Fig. 12. It is seen that ConnCrack can identify the cracks with high accuracy. Table 3 compares the results from the proposed method with other methods. The results from all other methods are reported in their papers except ResNet152-FCN [27] and VGG19-FCN [13] which are implemented by ourselves with the same learning rate as ConnCrack and pretraining on ImageNet. We can see that the proposed method outperforms other methods on CFD dataset in terms of precision and F1 score with large margin.

Table 3 – Comparison of performance for different methods on the CFD dataset

| **Method** | **Precision** | **Recall** | **F1 Score** |
|---|---|---|---|
| Canny[20] | 12.23% | 22.15% | 15.76% |
| CrackTree[20] | 73.22% | 76.45% | 70.80% |
| FFA[45] | 78.56% | 68.43% | 73.15% |
| CrackForest[20] | 82.28% | 89.44% | 85.71% |
| MFCD[45] | 89.90% | 89.47% | 88.04% |
| ResNet152-FCN | 87.83% | 88.19% | 88.01% |
| VGG19-FCN | 92.80% | 85.49% | 88.53% |
| CrackNet-V [3] | 92.58% | 86.03% | 89.18% |
| **ConnCrack** | **96.79%** | **87.75%** | **91.96%** |

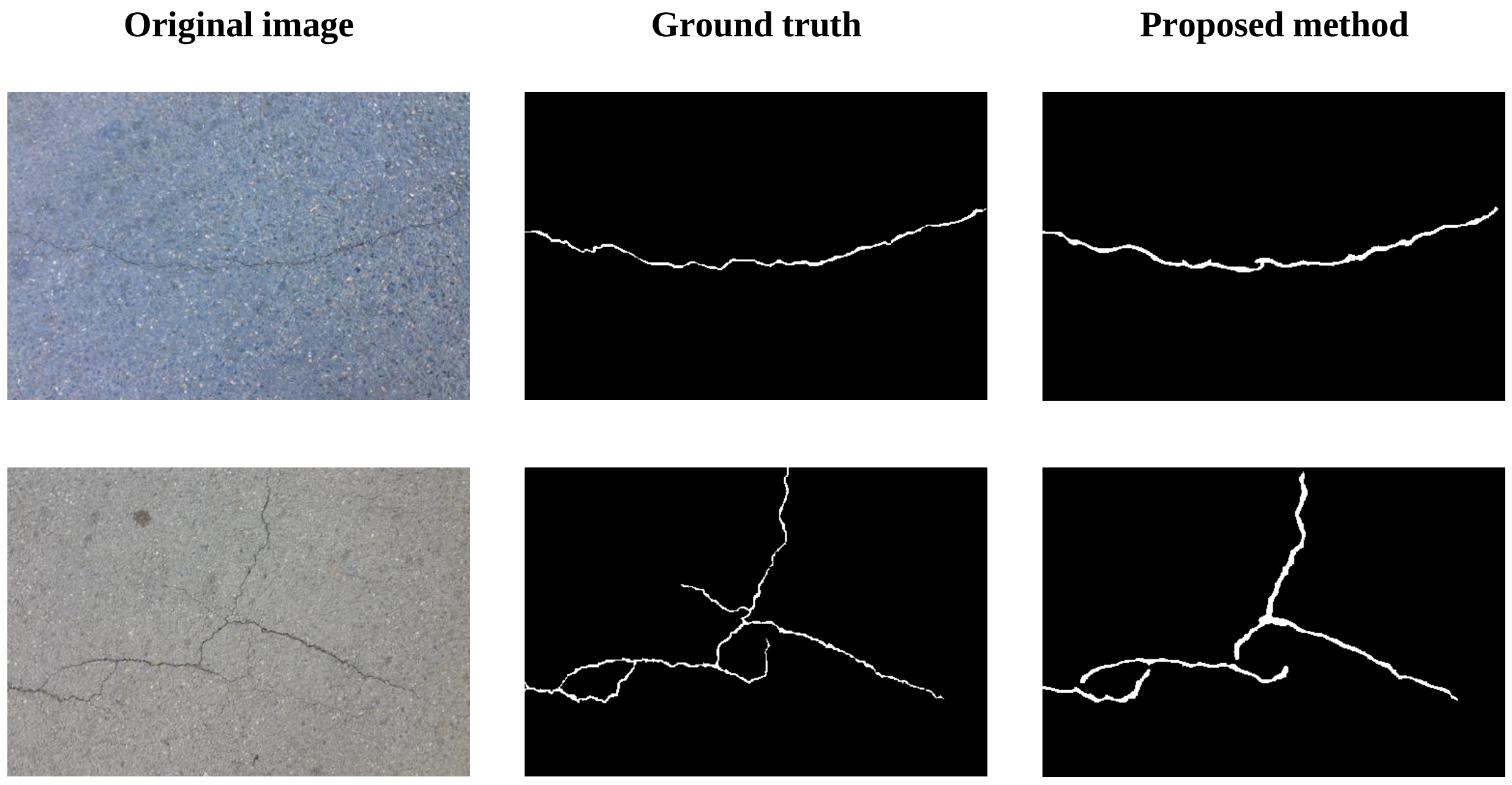

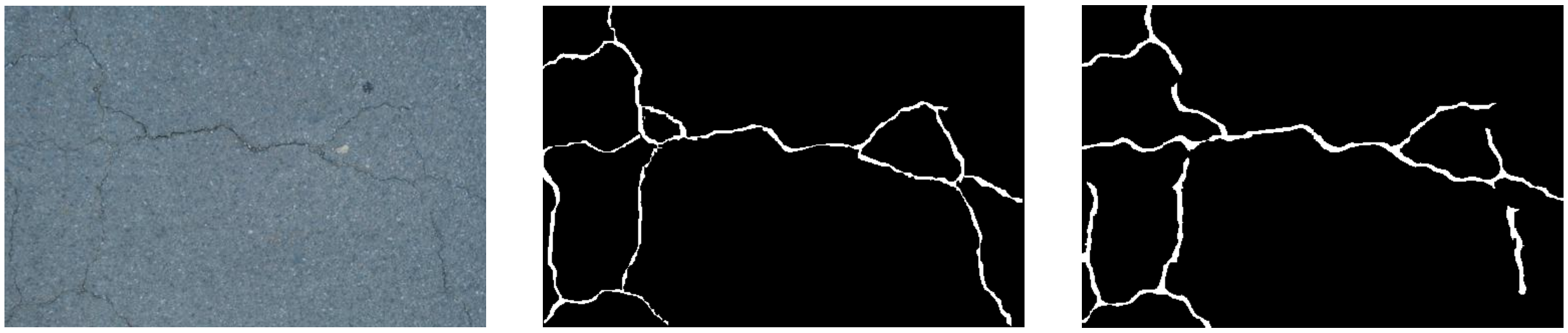

Fig. 12. Sample results for the CFD dataset

*5.2. Performance on EdmCrack600 dataset*

After pretraining on ImageNet and CFD datasets, the proposed method, ConnCrack, is further trained and tested on EdmCrack600 dataset. The images in EdmCrack600 are shuffled and split into 420/60/120 for training, validating and testing purposes. Similar to the pretraining, the images are first split into 256×256 patches, and are then integrated. The learning rate for the training is $1\times10^{-5}$, and $\lambda$ is set to $5\times10^{-6}$. The losses for training and validation sets are presented in Fig. 13. In the figures, as the training proceeds, we can see the content loss for generator barely reduces, but the cWGAN loss decreases. This demonstrates the superior training performance of the proposed method than traditional encoder-decoder networks because there is an additional source for weight updating. The discriminator loss increases as the training goes on because the predictions output by the generator become more difficult to distinguish.

The performance of the proposed method in terms of precision, recall and F1 score is presented in Table 4. The Sobel and Canny detectors are standard edge detection techniques [46]. CrackIT was proposed by Oliveira and Correia [33, 47] using a series of image processing techniques. ResNet152-FCN [27] and VGG19-FCN [13] created encoder-decoder networks as suggested by [38] with ResNet152 and VGG19 as backbone networks, respectively. U-Net was introduced by [48] for crack detection. All seven methods were tested on a desktop with Intel 8700k CPU, 32GB memory and Nvidia Titan V GPU with 5120 CUDA cores where Canny, Sobel and CrackIT methods were run on CPU and the other 4 deep learning based method were run on GPU. We can see in the table the proposed method outperforms other methods including other deep learning-based methods with large margin.

Some sample results from the proposed method and existing methods are presented in Fig. 14. We can see that rule-based methods cannot tackle with such complex situations where the cracks are mixed with illumination changes, shadows of trees, etc. The deep learning-based methods perform significantly better. In these methods, the illumination change and the texture of the pavement surfaces are not identified as cracks. However, ResNet152-FCN, VGG19-FCN and U-Net which utilize binary crack mask generates scattered output as described in section 4.2. Also, the noise appears at different locations in the results from those three methods. The proposed method overcomes the abovementioned issues using connectivity maps and DFS based thresholding, which results in more than 5% improvement in terms of F1 score. Regarding computational efficiency, the proposed method is slightly slower than VGG19-FCN but faster than ResNet152-FCN and U-Net.

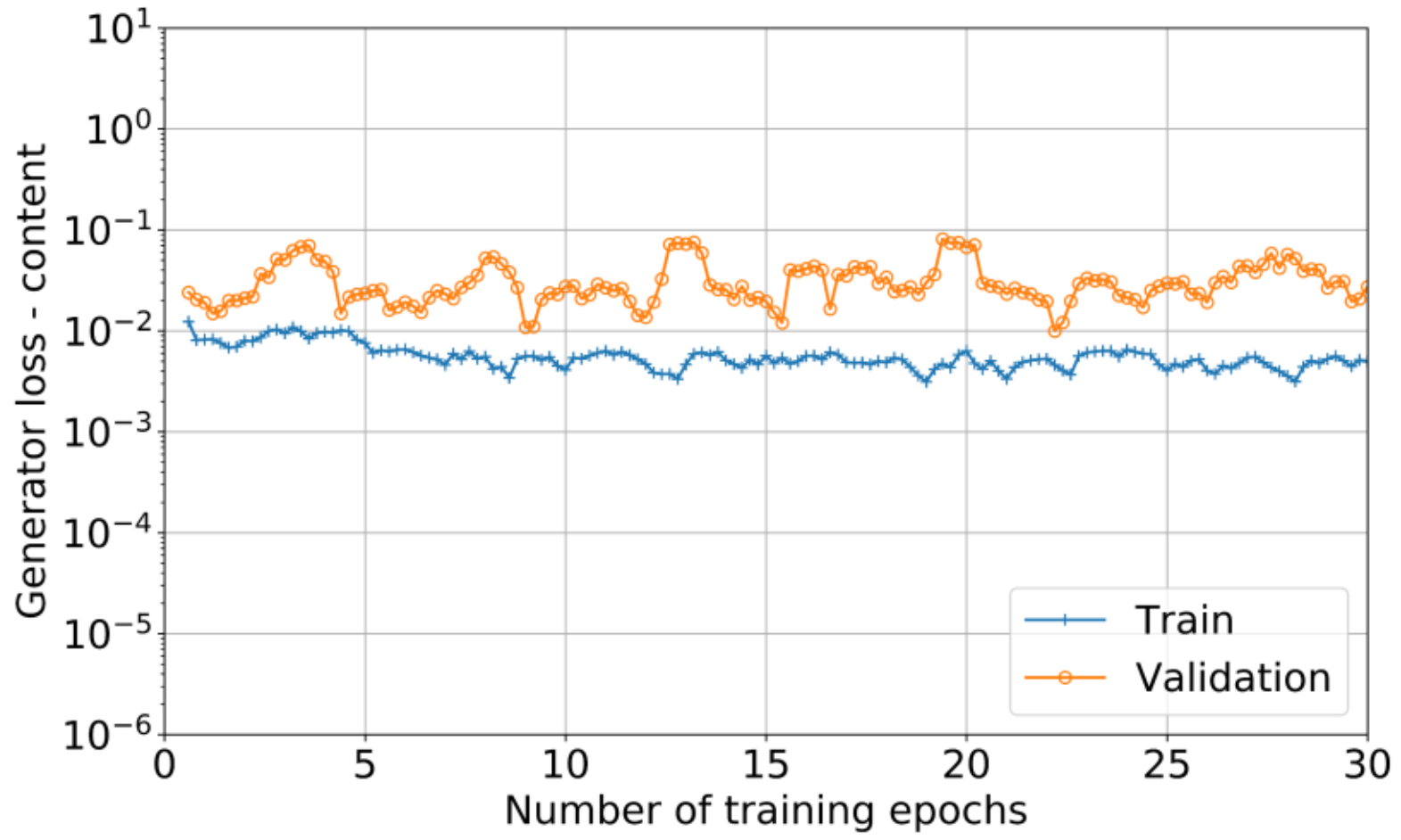

(a) Content loss for generator

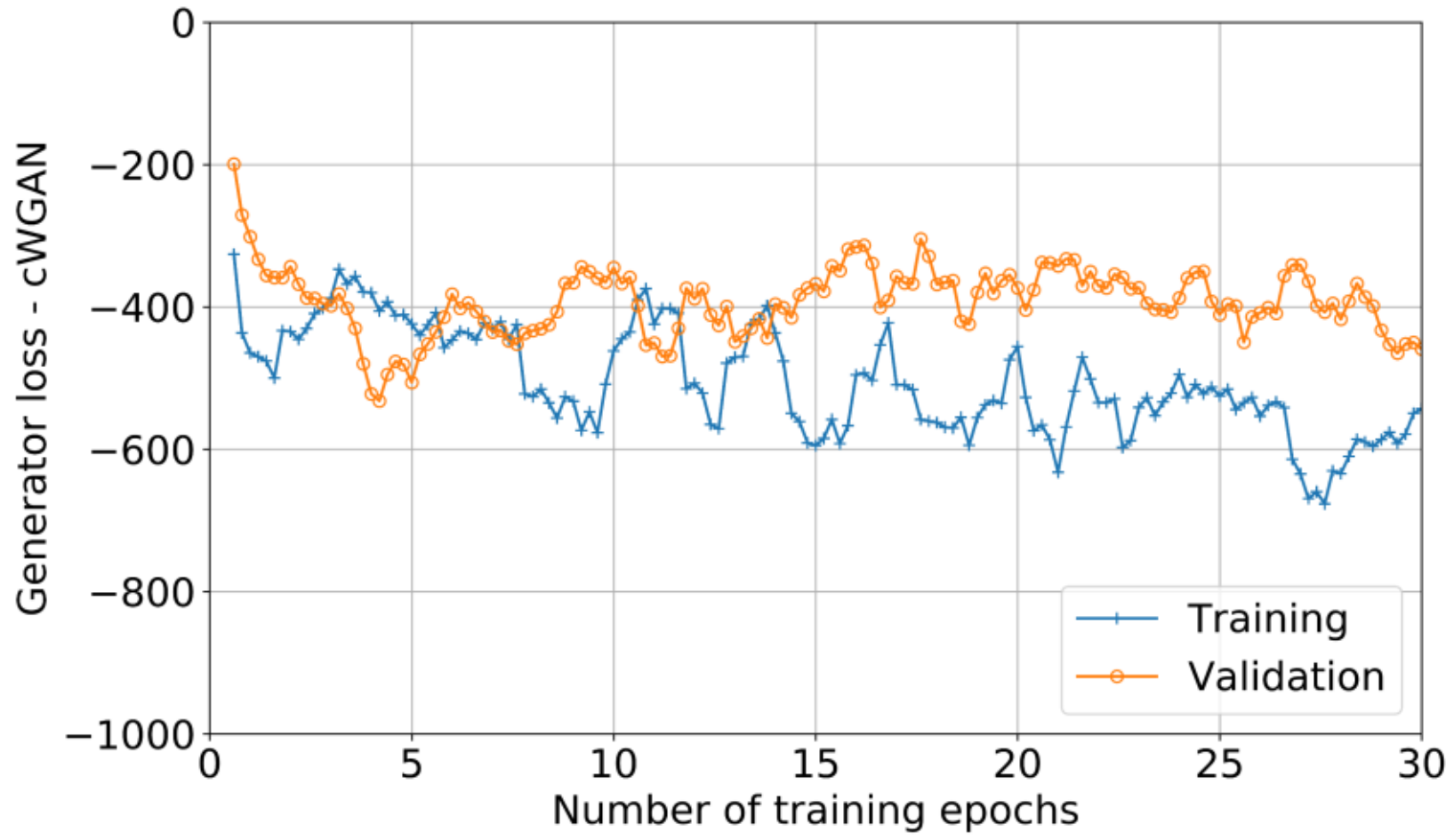

(b) cWGAN loss for generator

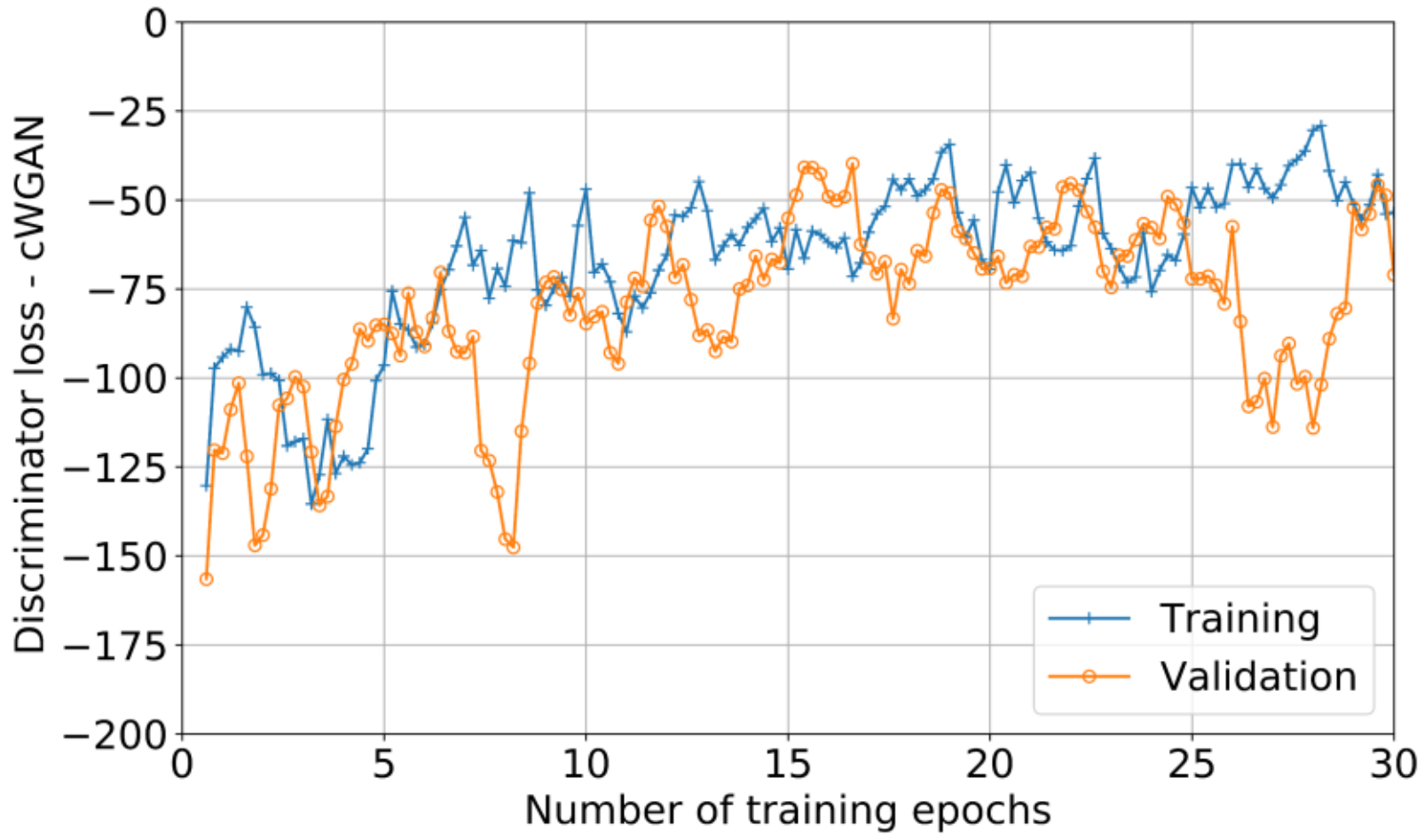

(c) cWGAN loss for discriminator

Fig. 13. Losses for EdmCrack600 dataset

Table 4 – Comparison of performance for different methods on the EdmCrack600 dataset

| **Method** | **Precision** | **Recall** | **F1 Score** | **Efficiency (sec/image)** |
|---|---|---|---|---|
| Canny | 1.69% | 34.17% | 3.14% | 0.12 |
| Sobel | 3.00% | 15.24% | 4.66% | **0.04** |
| CrackIT | 12.33% | 7.14% | 4.75% | 6.71 |
| ResNet152-FCN | 78.98% | 56.51% | 62.78% | 1.94 |
| VGG19-FCN | 80.22% | 59.93% | 65.18% | 1.33 |
| U-Net | 76.33% | 70.88% | 71.52% | 2.58 |
| **ConnCrack** | **80.88%** | **76.64%** | **76.98%** | 1.56 |

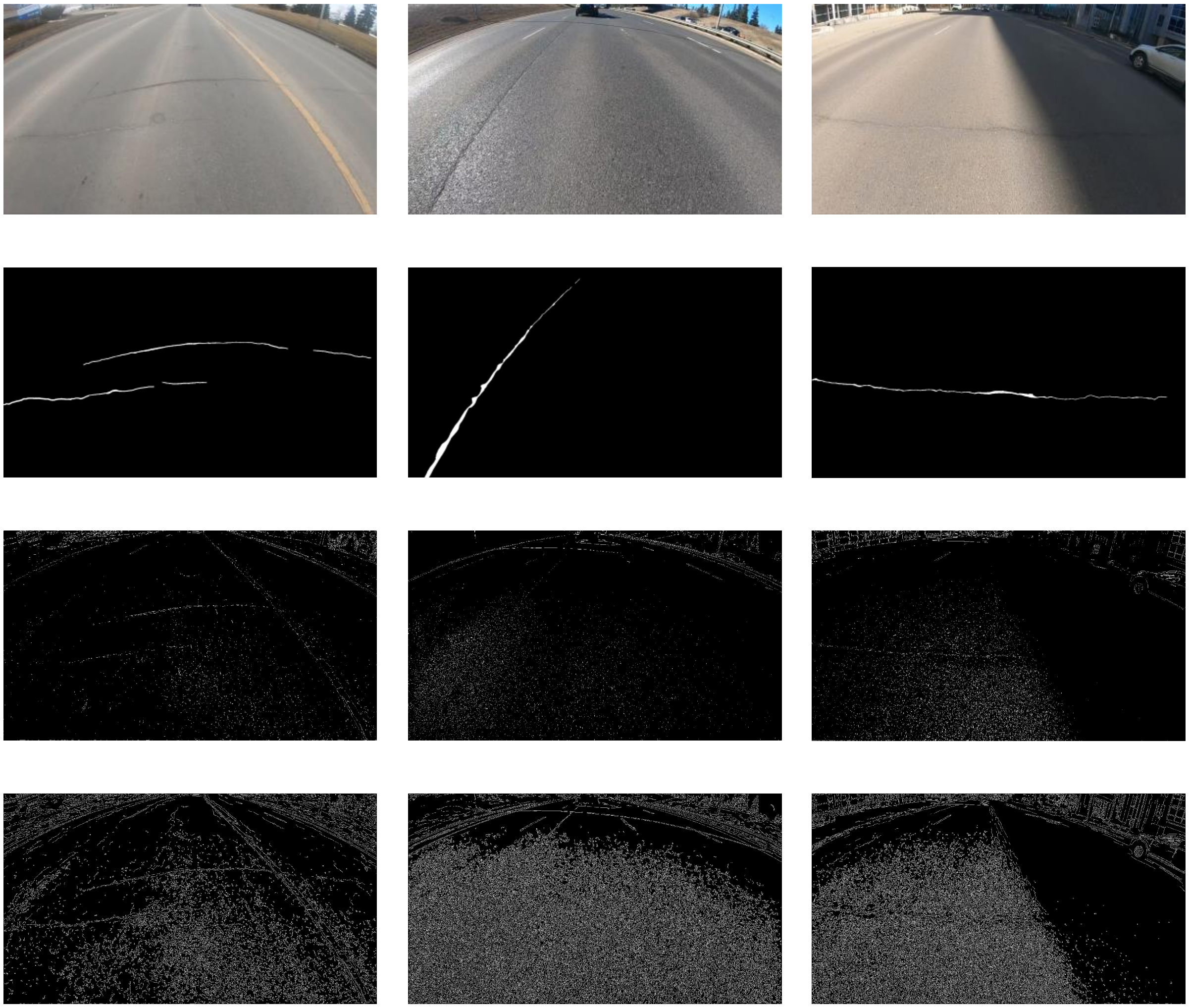

Original image

Ground truth

Sobel

Canny

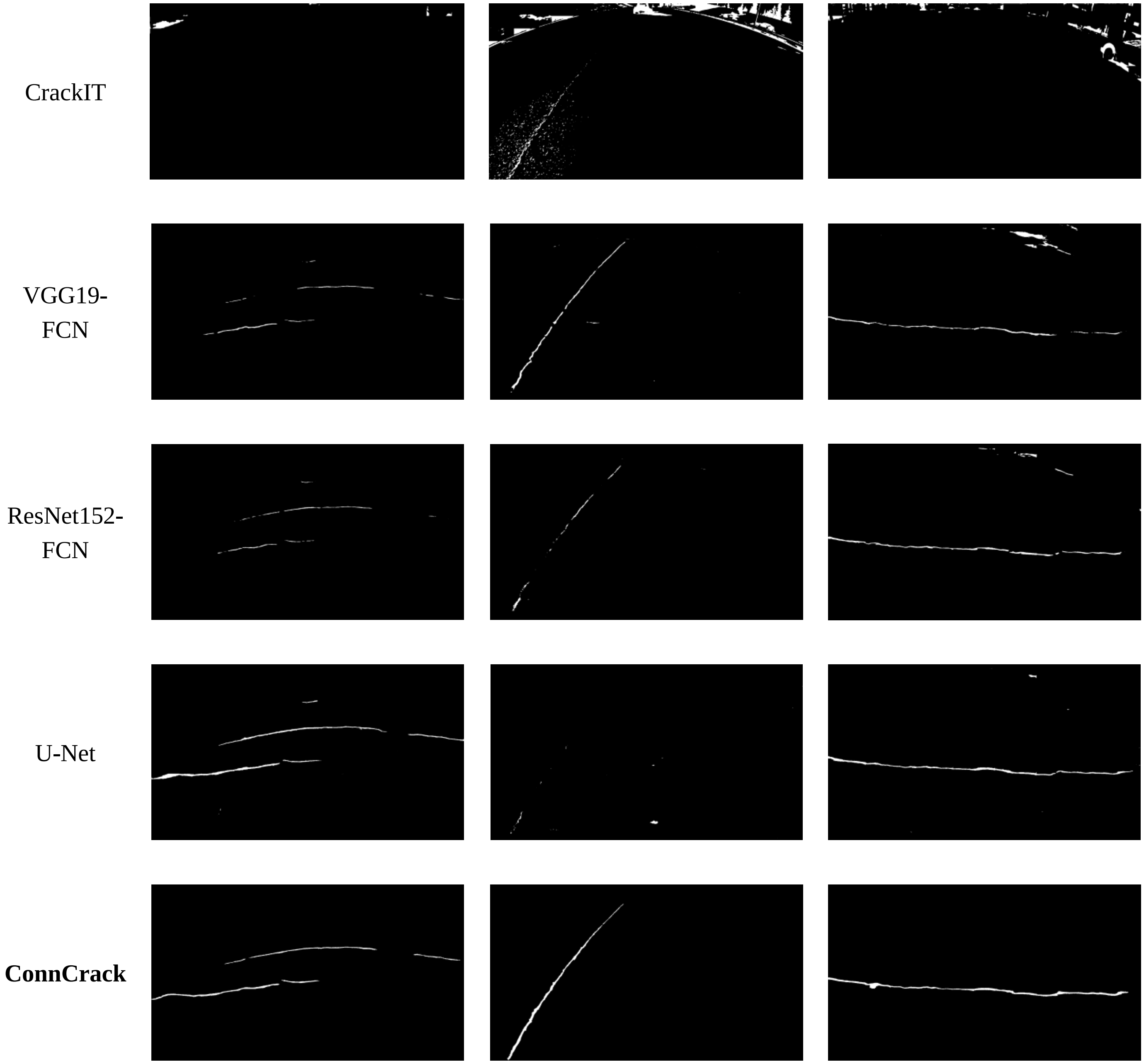

Fig. 14. Sample images and corresponding results for EdmCrack600 dataset

Although ConnCrack has achieved superior performance than other state-of-the-art methods, it still has difficulty in identifying the cracks correctly in some images. Fig. 15 shows two examples of wrongly identified images. In the left plots, the crack at the bottom was not identified by the ConnCrack. Looking at the original image, it is seen that the bottom crack is relatively blurry than other parts. This could be the reason that the proposed method cannot identify it properly. In the right plots, there are a lot of shadows from trees on the road surface. Although ConnCrack can identify the long and thick crack correctly, it also misidentifies some of the shadows as cracks. One possible solution to these issues is to collect more data with such critical cases.

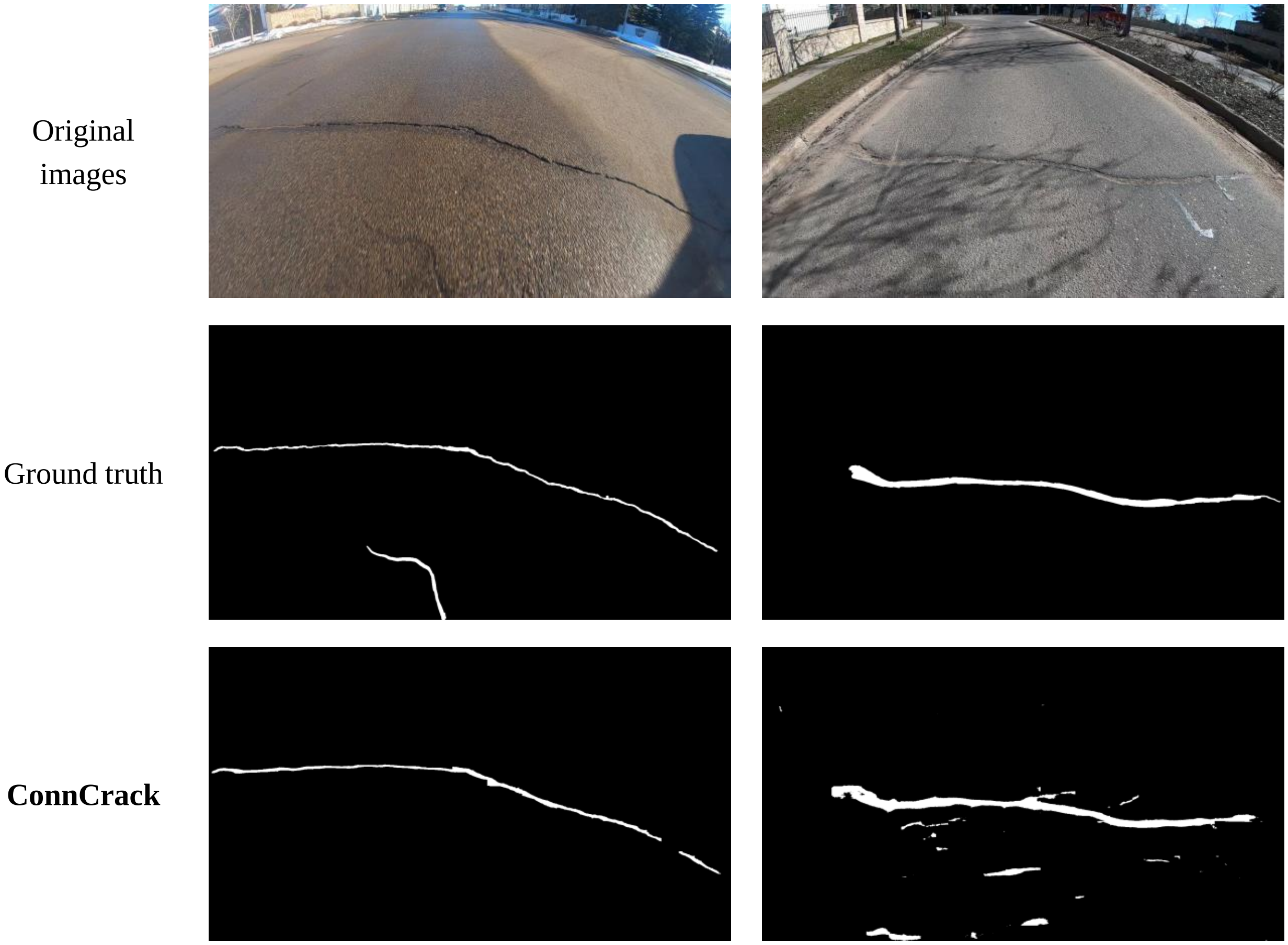

Fig. 15. Two wrongly identified images

In the dataset, the images are taken in perspective view. The parts that are farther from the center of the image have lower spatial resolutions (as explained in section 3). In this study, the perspective is not taken into consideration during the training and testing process, but it is meaningful to know how the perspective view affects the performance of the proposed method. In Fig. 15, all 120 images with 1920×1080 pixels in the test set are split into 16×9 grids. The precision, recall and F1 score are calculated for each small region separately for all 120 test images. The heat maps are generated for all three metrics where red means 100% and blue stands for 0%. The gray color represents no existence of cracks in that area. Looking at the Fig. 15(a), there is no significant difference in different regions in terms of precision except the top left corner. This means the precision is not very sensitive to the spatial resolution of the image. However, Fig. 15(b) shows that the recall is more sensitive to the location. The parts that are closer to the edges and corners have lower recall, which means the number of false negative pixels is higher in these regions. This shows that the proposed method is unlikely to predict the pixels that are too far from the centerline as cracks. This is because of the distortion and low resolution at the edges of images. As a combination of precision and recall, the F1 score has similar pattern as recall (see Fig. 15(c)).

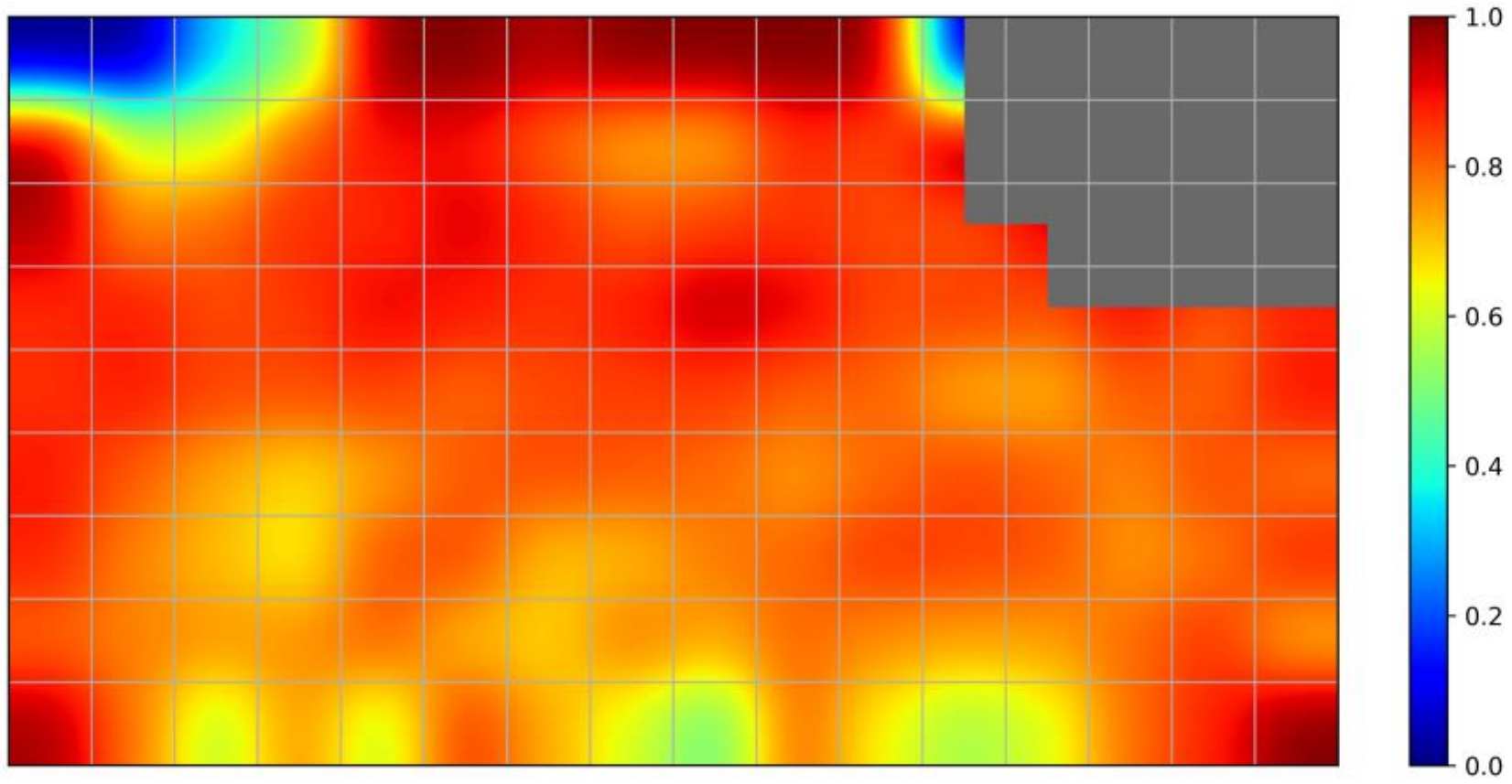

(a) Precision for different regions

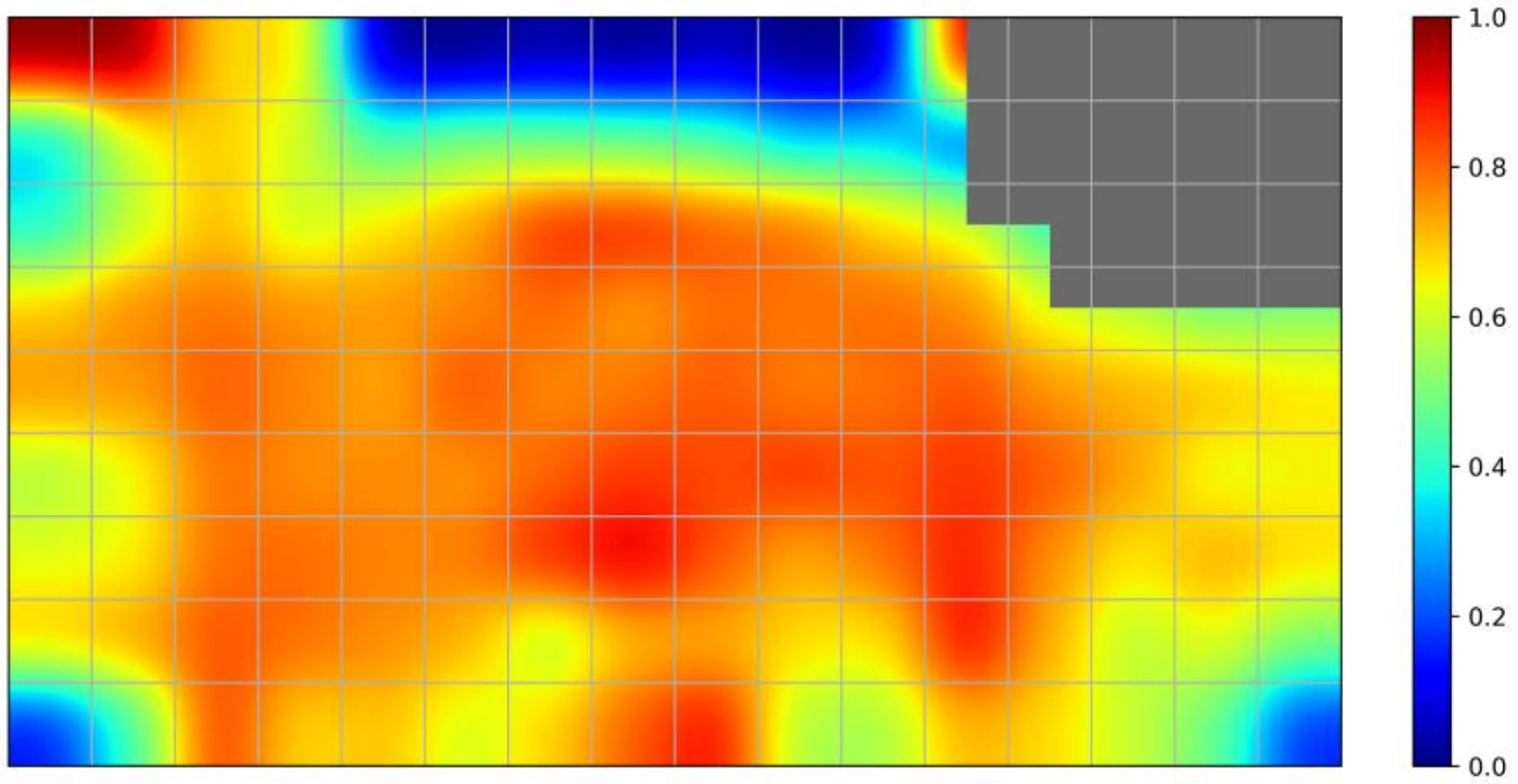

(b) Recall for different regions

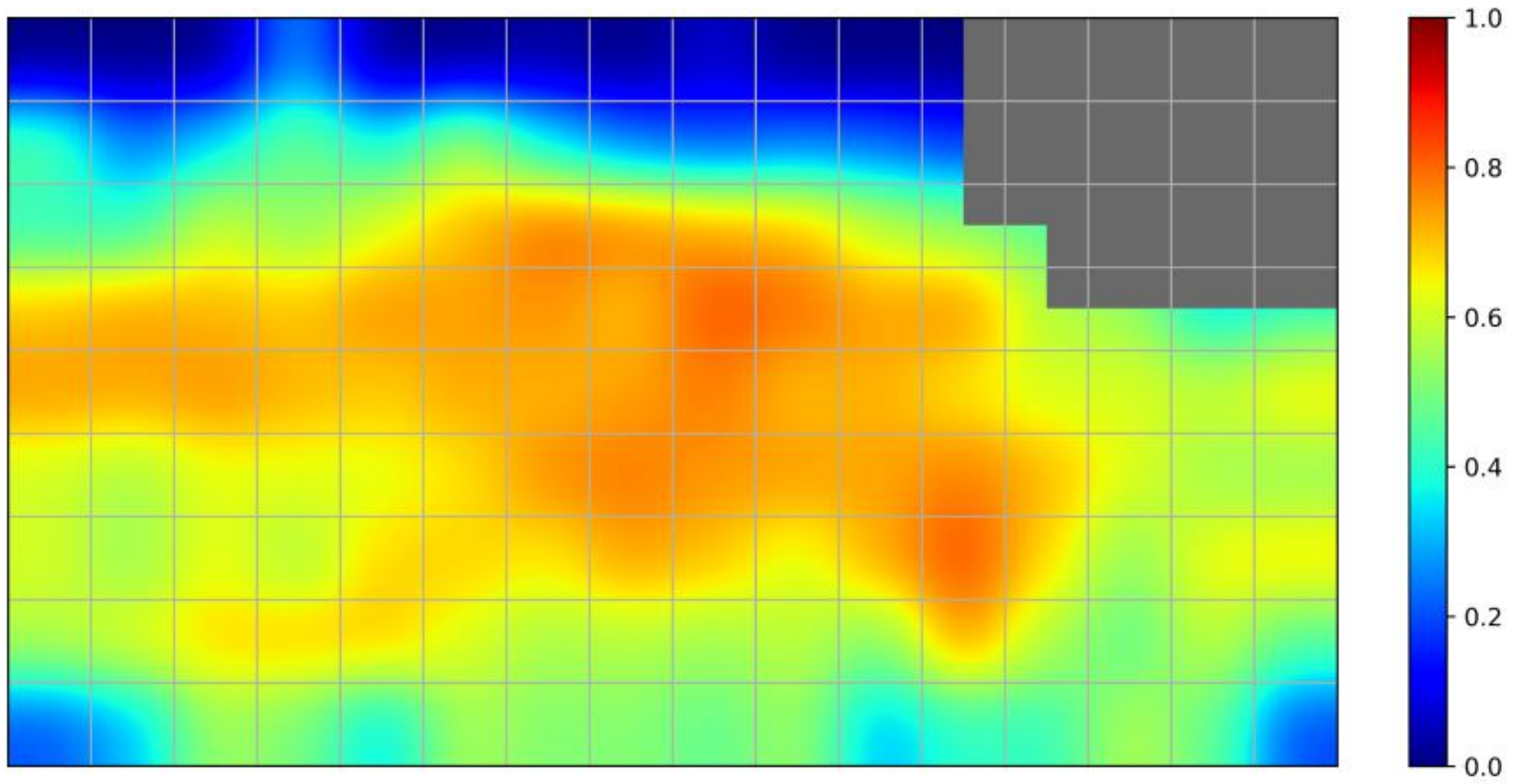

(c) F1 score for different regions

Fig.15. Performance of different regions of EdmCrack600 dataset

## *6. Conclusions*

In this paper, a cost effective road inspection solution using commercial grade sport camera and deep neural network is introduced. A deep learning-based algorithm called ConnCrack is proposed combining cWGAN and connectivity maps. The proposed method is first pretrained on ImageNet [44] and CFD dataset [20], and then trained and tested on EdmCrack600 dataset collected through our introduced road inspection solution. The following conclusions are drawn from this study:

1) Commercial grade sport cameras are feasible for road crack inspection.
2) The proposed method, ConnCrack, can outperform other existing methods on both publicly available dataset and our collected data.

Despite the success of the proposed method in this study, there are still limitations that needs to be addressed. For instance, the current version of the proposed method can only be used for crack detection. In the future, we will improve this method to detect multiple defects on the road simultaneously. Also, we will investigate more complex situations such as images taken at poor light condition and taken by low-speed camera with more significant blur issues. In the future, the research can be extended to back up camera in vehicles, and other sensors like accelerometer or GPS could be fused with the camera to provide further information about road conditions.

## *Acknowledgments*

This work was financially supported by the Natural Sciences and Engineering Research (NSERC) Discovery Grant.

## *References*

1. Chambon, S. and J.-M. Moliard, *Automatic road pavement assessment with image processing: review and comparison.* International Journal of Geophysics, 2011. **2011**.

2. Mohan, A. and S. Poobal, *Crack detection using image processing: A critical review and analysis.* Alexandria Engineering Journal, 2018. **57**(2): p. 787-798.

3. Fei, Y., et al., *Pixel-Level Cracking Detection on 3D Asphalt Pavement Images Through Deep-Learning-Based CrackNet-V.* IEEE Transactions on Intelligent Transportation Systems, 2019.

4. Dung, C.V. and L.D. Anh, *Autonomous concrete crack detection using deep fully convolutional neural network.* Automation in Construction, 2019. **99**: p. 52-58.

5. Vavrik, W., et al., *PCR evaluation: considering transition from manual to semi-automated pavement distress collection and analysis.* 2013.

6. Koch, C., et al., *A review on computer vision based defect detection and condition assessment of concrete and asphalt civil infrastructure.* Advanced Engineering Informatics, 2015. **29**(2): p. 196-210.

7. Lecompte, D., J. Vantomme, and H. Sol, *Crack detection in a concrete beam using two different camera techniques.* Structural Health Monitoring, 2006. **5**(1): p. 59-68.

8. Abdel-Qader, L., O. Abudayyeh, and M.E. Kelly, *Analysis of edge-detection techniques for crack identification in bridges.* Journal of Computing in Civil Engineering, 2003. **17**(4): p. 255-263.

9. Girshick, R., et al. *Rich feature hierarchies for accurate object detection and semantic segmentation*. in *Proceedings of the IEEE conference on computer vision and pattern recognition*. 2014.

10. Ren, S., et al. *Faster r-cnn: Towards real-time object detection with region proposal networks*. in *Advances in neural information processing systems*. 2015.

11. Redmon, J. and A. Farhadi, *Yolov3: An incremental improvement.* arXiv preprint arXiv:1804.02767, 2018.

12. Chen, F.-C. and M.R. Jahanshahi, *NB-CNN: deep learning-based crack detection using convolutional neural network and naive Bayes data fusion.* IEEE Transactions on Industrial Electronics, 2018. **65**(5): p. 4392-4400.

13. Yang, X., et al., *Automatic pixel-level crack detection and measurement using fully convolutional network.* Computer-Aided Civil and Infrastructure Engineering, 2018. **33**(12): p. 1090-1109.

14. Mei;, Q. and M. Gül;. *EdmCrack600*. 2019 [cited 2019 July 10]; Available from: https://github.com/mqp2259/EdmCrack600.

15. Gavilán, M., et al., *Adaptive road crack detection system by pavement classification.* Sensors, 2011. **11**(10): p. 9628-9657.

16. Zou, Q., et al., *CrackTree: Automatic crack detection from pavement images.* Pattern Recognition Letters, 2012. **33**(3): p. 227-238.

17. Amhaz, R., et al., *Automatic crack detection on two-dimensional pavement images: An algorithm based on minimal path selection.* IEEE Transactions on Intelligent Transportation Systems, 2016. **17**(10): p. 2718-2729.

18. Hu, Y., C.-x. Zhao, and H.-n. Wang, *Automatic pavement crack detection using texture and shape descriptors.* IETE Technical Review, 2010. **27**(5): p. 398-405.

19. Mathavan, S., M. Rahman, and K. Kamal, *Use of a self-organizing map for crack detection in highly textured pavement images.* Journal of Infrastructure Systems, 2014. **21**(3): p. 04014052.

20. Shi, Y., et al., *Automatic road crack detection using random structured forests.* IEEE Transactions on Intelligent Transportation Systems, 2016. **17**(12): p. 3434-3445.

21. He, K., et al. *Mask r-cnn*. in *Proceedings of the IEEE international conference on computer vision*. 2017.

22. Zhang, L., et al. *Road crack detection using deep convolutional neural network*. in *Image Processing (ICIP), 2016 IEEE International Conference on*. 2016. IEEE.

23. Cha, Y.J., W. Choi, and O. Büyüköztürk, *Deep learning-based crack damage detection using convolutional neural networks.* Computer-Aided Civil and Infrastructure Engineering, 2017. **32**(5): p. 361-378.

24. Gopalakrishnan, K., et al., *Deep Convolutional Neural Networks with transfer learning for computer vision-based data-driven pavement distress detection.* Construction and Building Materials, 2017. **157**: p. 322-330.

25. Nhat-Duc, H., Q.-L. Nguyen, and V.-D. Tran, *Automatic recognition of asphalt pavement cracks using metaheuristic optimized edge detection algorithms and convolution neural network.* Automation in Construction, 2018. **94**: p. 203-213.

26. Ni, F., J. Zhang, and Z. Chen, *Pixel-level crack delineation in images with convolutional feature fusion.* Structural Control and Health Monitoring, 2019: p. e2286.

27. Bang, S., et al., *Encoder–decoder network for pixel-level road crack detection in black-box images.* Computer-Aided Civil and Infrastructure Engineering, 2019.

28. Maeda, H., et al., *Road damage detection using deep neural networks with images captured through a smartphone.* arXiv preprint arXiv:1801.09454, 2018.

29. Saranga, D. *Kia Sorento (2013)*. 2019 [cited 2019 July 10]; Available from: https://www.the-blueprints.com/blueprints/cars/kia/56201/view/kia_sorento_2013/.

30. Eisenbach, M., et al. *How to get pavement distress detection ready for deep learning? A systematic approach*. in *2017 international joint conference on neural networks (IJCNN)*. 2017. IEEE.

31. Yang, F., et al., *Feature Pyramid and Hierarchical Boosting Network for Pavement Crack Detection.* arXiv preprint arXiv:1901.06340, 2019.

32. Eisenbach, M., et al. *How to get pavement distress detection ready for deep learning? A systematic approach*. in *2017 International Joint Conference on Neural Networks (IJCNN)*. 2017.

33. Oliveira, H. and P.L. Correia. *CrackIT — An image processing toolbox for crack detection and characterization*. in *2014 IEEE International Conference on Image Processing (ICIP)*. 2014.

34. Goodfellow, I., et al. *Generative adversarial nets*. in *Advances in neural information processing systems*. 2014.

35. Kampffmeyer, M., et al., *ConnNet: A Long-Range Relation-Aware Pixel-Connectivity Network for Salient Segmentation.* IEEE Transactions on Image Processing, 2019. **28**(5): p. 2518-2529.

36. Huang, G., et al. *Densely connected convolutional networks*. in *Proceedings of the IEEE conference on computer vision and pattern recognition*. 2017.

37. LeCun, Y., et al., *Gradient-based learning applied to document recognition.* Proceedings of the IEEE, 1998. **86**(11): p. 2278-2324.

38. Long, J., E. Shelhamer, and T. Darrell. *Fully convolutional networks for semantic segmentation*. in *Proceedings of the IEEE conference on computer vision and pattern recognition*. 2015.

39. Isola, P., et al. *Image-to-image translation with conditional adversarial networks*. in *Proceedings of the IEEE conference on computer vision and pattern recognition*. 2017.

40. Mirza, M. and S. Osindero, *Conditional generative adversarial nets.* arXiv preprint arXiv:1411.1784, 2014.

41. Arjovsky, M., S. Chintala, and L. Bottou, *Wasserstein gan.* arXiv preprint arXiv:1701.07875, 2017.

42. Radford, A., L. Metz, and S. Chintala, *Unsupervised representation learning with deep convolutional generative adversarial networks.* arXiv preprint arXiv:1511.06434, 2015.

43. Pan, S.J. and Q. Yang, *A survey on transfer learning.* IEEE Transactions on knowledge and data engineering, 2010. **22**(10): p. 1345-1359.

44. Deng, J., et al. *Imagenet: A large-scale hierarchical image database*. in *Computer Vision and Pattern Recognition, 2009. CVPR 2009. IEEE Conference on*. 2009. Ieee.

45. Li, H., et al., *Automatic Pavement Crack Detection by Multi-Scale Image Fusion.* IEEE Transactions on Intelligent Transportation Systems, 2018(99): p. 1-12.

46. Gonzalez, R.C. and P. Wintz, *Digital image processing*. Reading, Mass., Addison-Wesley Publishing Co., Inc.(Applied Mathematics and Computation. 1987. 451.

47. Oliveira, H. and P.L. Correia, *Automatic road crack detection and characterization.* IEEE Transactions on Intelligent Transportation Systems, 2012. **14**(1): p. 155-168.

48. Liu, Z., et al., *Computer vision-based concrete crack detection using U-net fully convolutional networks.* Automation in Construction, 2019. **104**: p. 129-139.

# List of Tables

# List of Figures